\documentclass{article}

\usepackage{PRIMEarxiv}

\usepackage[hyphens]{url}
\usepackage{lineno,hyperref}
\usepackage[numbers]{natbib}
\modulolinenumbers[5]

\usepackage{amsmath,amsfonts,bm}
\usepackage{amsthm}
\usepackage{mathtools}
\usepackage{bigdelim}
\usepackage{bbm}
\usepackage{amsfonts}
\usepackage{graphicx}
\usepackage{multirow}
\usepackage{booktabs}
\usepackage{float}
\usepackage{tabularx}
\usepackage{xcolor}
\usepackage{longtable}
\usepackage{subcaption}
\usepackage{pdflscape}
\usepackage{afterpage}
\usepackage{tikz}
\usepackage{wrapfig}
\usepackage{todonotes}
\usetikzlibrary{matrix,chains,positioning,decorations.pathreplacing,arrows}
\usepackage{fancyhdr}   
\usepackage{array}
\newcommand{\PreserveBackslash}[1]{\let\temp=\\#1\let\\=\temp}
\newcolumntype{C}[1]{>{\PreserveBackslash\centering}p{#1}}

\newcommand{\zit}[1]{z_{i,#1}}

\newcommand{\vxit}[1]{\mathbf{x}_{i,#1}}

\newcommand{\vzit}[1]{\mathbf{z}_{i,#1}}

\newcommand{\vXit}[1]{\mathbf{X}_{i,#1}}

\newcommand{\vXt}[1]{\mathbf{X}_{#1}}
\newcommand{\vZt}[1]{\mathbf{Z}_{#1}}

\newcommand{\bR}{\mathbb{R}}

\newcommand{\deep}{deep }
\newcommand{\Deep}{Deep } 

\newcommand{\gl}{global-local }
\newcommand{\Gl}{Global-local }

\newcommand{\defeq}{\vcentcolon=}

\newcommand{\owner}[1]{}
\renewcommand{\owner}[1]{\colorbox{yellow}{\textbf{{#1}}}}

\graphicspath{ {figures/} }
\title{Deep Learning for Time Series Forecasting: Tutorial and Literature Survey
}

\newcommand*\samethanks[1][\value{footnote}]{\footnotemark[#1]}

\author{
  Konstantinos Benidis\thanks{Equal contribution.} \\
  Amazon Research \\
  Berlin, Germany\\
  \texttt{kbenidis@amazon.com} \\
   \And
   Syama Sundar Rangapuram \\
   Amazon Research \\
  Berlin, Germany\\
  \texttt{rangapur@amazon.com} \\
     \And
  Valentin Flunkert \\
  Amazon Research \\
  Berlin, Germany\\
  \texttt{flunkert@amazon.com} \\
  \And
Yuyang Wang\\
  Amazon Research \\
  East Palo Alto, CA, USA\\
  \texttt{yuyawang@amazon.com} \\
  \And
Danielle Maddix\\
  Amazon Research \\
  East Palo Alto, CA, USA\\
  \texttt{dmmaddix@amazon.com} \\
  \And
Caner Turkmen\\
  Amazon Research \\
  Berlin, Germany\\
  \texttt{atturkm@amazon.com} \\
  \And
Jan Gasthaus \\
  Amazon Research \\
  Berlin, Germany\\
  \texttt{gasthaus@amazon.com} \\
  \And
Michael Bohlke-Schneider \\
  Amazon Research \\
    Berlin, Germany\\
  \texttt{bohlkem@amazon.com} \\
  \And
David Salinas \\
  Amazon Research \\
    Berlin, Germany\\
  \texttt{dsalina@amazon.com} \\
  \And
Lorenzo Stella\\
  Amazon Research \\
    Berlin, Germany\\
  \texttt{stellalo@amazon.com} \\
  \And
Fran{\c{c}}ois-Xavier Aubet\\
  Amazon Research \\
    Berlin, Germany\\
  \texttt{aubetf@amazon.com} \\
  \And
Laurent Callot\\
  Amazon Research \\
    Berlin, Germany\\
  \texttt{lcallot@amazon.com} \\
  \And
Tim Januschowski\samethanks \hspace{0.5mm} \thanks{Work done while at AWS.}\\
  Zalando SE \\
    Berlin, Germany\\
  \texttt{ tim.januschowski@zalando.de} \\
}

\date{}

\begin{document}

\maketitle

\begin{abstract}
  Deep learning based forecasting methods have 
  become the methods of choice in many applications of time series prediction or \emph{forecasting} often outperforming other approaches. Consequently, over the last years, these methods are now ubiquitous in large-scale industrial forecasting applications and have consistently ranked among the best entries in forecasting competitions (e.g.,\ M4 and M5).
  This practical success has further increased the academic interest to understand and improve \deep forecasting methods.
  In this article we provide an introduction and overview of the field:
  We present important building blocks for \deep forecasting in some depth; using these building blocks, we then survey the breadth of the recent \deep forecasting literature.
\end{abstract}


\section{Introduction}
\label{sec:intro}

Forecasting is the task of extrapolating time series into the future. It has many important applications~\cite{petropoulos2020forecasting} such as
forecasting the demand for items sold by retailers~\citep{Croston1972, wen2017multi, salinas2020deepar, mukherjee2018armdn, kasun19, bose2017probabilistic},
the flow of traffic~\citep{laptev2017,lv2014traffic,li2018diffusion}, 
the demand and supply of energy~\citep{DIMOULKAS2019,SMYL2019,LI2019, Saxena2019},
or the covariance matrix, volatility and long-tail distributions in finance~\cite{callot2017modeling,callot2019nodewise, volatility,Ballestra2019, neurips_18_long_tail}. As such, it is a well-studied area (e.g., see~\cite{hyndman2018forecasting} for an introduction) with its own dedicated research community. The machine learning, data science, systems, and operations research communities as well as application-specific research communities have also studied the problem intensively (e.g., see a series of recent tutorials~\cite{faloutsos19forecasting2,faloutsos2018forecasting,Faloutsos2019,faloutsos20}).
In contrast to traditional forecasting applications,  modern incarnations often exhibit large panels of related time series, all of which need to be forecasted simultaneously~\cite{janusch18}. Although these problem characteristics make them amenable to deep learning or neural networks (NNs), as in many other domains over the course of history, NNs were not always a standard tool to tackle such problems. Indeed, their effectiveness has historically been regarded as mixed (e.g.,~\cite{zhang1998forecasting}).

The history of NNs starts in 1957~\citep{rosenblatt1957perceptron} and in 1964 for NNs in forecasting~\citep{hu1964}. 
Since then, interest in NNs has oscillated, with upsurges in attention 
attributable to breakthroughs.  The application of NNs in time series forecasting has followed the general popularity, typically with a lag of a few years. 
Examples of such breakthroughs include \citet{rumelhart1985learning, rumelhart1986learning} that popularized the training of multilayer perceptrons (MLPs) using back-propagation. 
Significant advances were made subsequently 
such as
the use of convolutional  NNs (CNNs) \citep{lecun1995convolutional}, 
and Long Short Term Memory (LSTM) \citep{hochreiter1997long} cells that address the issue of recurrent  NNs' (RNNs) training, just to name a few. 
Despite these advances, NNs remained hard to train and difficult to work with. Methods such as Support Vector Machines (SVMs) \cite{boser1992svm} and Random Forests \citep{ho1995random} that were developed in the 1990s proved to be highly effective (\citet{lecun1995comparison} found that SVMs were as good as the best designed  NNs available at the time) and were supported by attractive 
theory. This shifted the interest of researchers away from NNs. Forecasting was no exception and results obtained with NNs were mostly mixed as reflected in a highly cited review~\citep{zhang1998forecasting}. 
The breakthrough that marked the dawn of the deep learning era came in 2006 when~\citet{hinton2006fast} showed that it was possible to train NNs with a large number of layers (deep) if the weights are initialized appropriately. Accordingly, deep learning has had a sizable impact on forecasting~\cite{langkvist2014review} and NNs have long entered the canon of standard techniques for forecasting~\citep{hyndman2018forecasting}. 
New models specifically designed for forecasting tasks have been proposed, 
taking advantage of deep learning to supercharge classical forecasting models or to develop entirely novel approaches. This recent burst of attention on \emph{\deep forecasting} models is the latest twist in a long and rich history. 

Driven by the availability of (closed-source) large time series panels, the potential of \deep forecasting models, i.e., forecasting models based on NNs, has been exploited primarily in applied industrial research divisions over the last years~\cite{laptev2017,gasthaus2019probabilistic,salinas2020deepar,wen2017multi}.\footnote{Forecasting is an example of a sub-discipline in the machine learning community where the comparatively modest attention it receives in published research is in stark contrast to a tremendous business impact.} 
With the overwhelming success of \deep forecasting methods in the M4 competition~\cite{smyl2018m4}, this has convinced also formerly skeptical academics~\cite{makridakis18,makridakis2018m4}. 
In the most recent M5 competition,  \deep forecasting methods were the second and third placed solutions \citep{makridakis2021m5} although the competition was otherwise dominated by tree-based forecasting methods such as LightGBM~\cite{LightGBM17} and XGBoost~\cite{Chen2016}, see e.g.,~\citep{januschowski21trees}. Modern software frameworks~\cite{abadi2016tensorflow,paszke2017automatic,chen2015mxnet} have sped up the development of NN models and dedicated forecasting packages available~\cite{alexandrov2019gluonts}. 

While the history of NNs for forecasting is rich, the focus of this article is on more recent developments in NN for forecasting, roughly since the time that the term ``deep learning'' was coined. As such, we do not attempt to give a complete historical overview and sacrifice comprehensiveness for recency. 
The main objectives of this article are to educate on, review and popularize the recent developments in forecasting driven by NNs for a general audience. Therefore, we place emphasis on an educational aspect via a tutorial of \deep forecasting in the first part (Section~\ref{sec:building_blocks}). In the second part, Section~\ref{sec:lit_review}, we provide an overview of the state-of-the-art of modern \deep forecasting models. Our exposition is driven by an attempt to identify the main building blocks of modern \deep forecasting models which hopefully enables the reader to digest the rapidly increasing literature more easily. We do not attempt a taxonomy of all existing methods and our selection of the building blocks is opinionated, motivated by our experience of innovating in this area with a strong focus on practical applicability. Compared with other surveys~\cite{Lim_2021,hewamalage2019recurrent,zhang1998forecasting}, we provide a more comprehensive overview with a particular focus on recent, advanced topics. 
Finally, in Section~\ref{sec:conclusions}, we conclude and speculate on potentially fruitful areas for future research.


\section{\Deep Forecasting: A Tutorial}
\label{sec:building_blocks}

In the following, we formalize the forecasting problem, summarize those advances in deep learning that we deem as the most relevant for forecasting, expose important building blocks for NNs and discuss archetypal models in detail. For general improvements that fueled the deep learning renaissance, like weight initialization, optimization algorithms or general-purpose components such as activation functions, we refer to standard textbooks like~\citep{GBC2016}. We are aware to be opinionated in both the selection of topics as well as the style of exposition. We attempt to take a perspective akin to a \deep forecasting model builder who would compose a forecasting model out of several building blocks such as NN architectures, input transformations and output representations. 
Although not all models will fit perfectly into this exposition, it is our hope that this downside is outweighed by the benefit of allowing the inclined reader to invent new models more easily.

\subsection{Notation and Formalization of the Forecasting Problem}
\label{ssec:formalization}

Matrices, vectors and scalars are denoted by uppercase bold, lowercase bold and lowercase normal letters, i.e., $\mathbf{X}$, $\mathbf{x}$ and $x$, respectively. 
Let $\mathcal{Z} = \{\vzit{1:T_i}\}_{i=1}^N$ 
be a set of $N$ univariate time series, where $\vzit{1:T_i} = (\zit{1}, \dots, \zit{T_i})$, $\zit{t}$ is the value of the $i$-th time series at time $t$ and $\vZt{t_1: t_2}$ the values of all $N$ time series at the time slice $[t_1,t_2]$. 
Typical examples for the domain of the time series values include $\mathbb{R}, \mathbb{N}, \mathbb{Z}, [0,1]$. 
The set of time series is associated with a set of covariate vectors denoted by $\mathcal{X} = \{\vXit{1:T_i}\}_{i=1}^N$, with $\vxit{t}\in\bR^{d_x}$. Note that each vector $\vxit{t}$ can include both time-varying or static features. We denote by $\alpha$ a general input in a model (that can be any combination of covariates and lagged values of the target) and by $\beta$ a general output. Since $\alpha$ and $\beta$ refer to a general case, we always represent them with lowercase normal letters. We denote by $\theta$ the parameters of a model (e.g., parameters of a distribution) and by $\Phi$ the learnable free parameters of the underlying NN (e.g., the weights and biases).

In the most general form, the object of interest in forecasting is the conditional distribution
\begin{equation}
\label{eq:general_set_up}
	p(\vZt{t+1:t+h} | \vZt{1:t}, \vXt{1:t+h}; \theta),
\end{equation}
where $\theta$ are the parameters of a (probabilistic) model. 
Eq.~\eqref{eq:general_set_up} is general in the sense that each $\mathbf{z}_i \in \mathcal{Z}$ is multidimensional (the length of the time series), 
$\mathcal{Z}$ is multivariate (the number of time series $|\mathcal{Z}| = N>1$) and the forecast is multi-step ($h$ steps). Varying degrees of simplification of Eq.~\eqref{eq:general_set_up} are considered in the literature, for example by assuming factorizations of $p$ and different ways of estimating $\theta$. In the following, we present the three archetypical models for addressing Eq.~\eqref{eq:general_set_up}.

\textbf{Local univariate model:} A separate (\emph{local}) model is trained independently for each of the $N$ time series, modelling the predictive distribution
\begin{equation}
\label{eq:local_univariate}
	p(\vzit{t+1:t+h} | \vzit{1:t}, \vXit{1:t+h} ; \theta_i),\quad \theta_i = \Psi(\vzit{1:t}, \vXit{1:t+h}),
\end{equation}
where $\Psi$ is a generic function mapping input features to the parameters $\theta_i$ of the probabilistic model that are local to the $i$-th time series. Note that one may use multidimensional covariates $\vxit{t}$ for each of the $N$ models, but they are still solving a univariate problem, i.e., forecasting only one time series. 
The use of covariates common to all $N$ models is possible but any pattern that is learned in one model is not used in another (unless provided explicitly which prohibits parallel training). Many classical approaches fall into this category and traditionally NNs were employed in this local fashion (e.g.,~\citep{zhang1998forecasting}).
Note that this approach is not suitable for cold start problems: i.e., forecasting a time series without historical values. 
	
\textbf{Global univariate model:} A single, \emph{global} model~\citep{januschowski19,montero2020principles} is trained using available data from all $N$ time series. However, the model is still used to predict a univariate target. 
It does not produce joint forecasts of all time series but forecasts of any single time series at a time. This is also sometimes referred to as a cross-learning approach, e.g.,~\citep{SEMENOGLOU20211072}.
In a more general form, global univariate models specialize Eq.~\eqref{eq:general_set_up} to
\begin{equation}
\label{eq:global_univariate}
p(\vzit{t+1:t+h} | \vZt{1:t}, \vXt{1:t+h}; \theta_i),\quad \theta_i = \Psi(\vzit{1:t}, \vXit{1:t+h},\Phi),
\end{equation}
where $\Phi$ are shared parameters among all $N$ time series.

In this article, $\Psi$ in global models is usually a NN and $\mathbf{X}_i$ include item-specific features to allow the model to distinguish between the time series. 
Although the parameters $\theta_i$ of the probabilistic model for each time series are different, they are still predicted using shared parameters (or weights) $\Phi$ in $\Psi$.		
This allows for efficient learning since the model pools information from all time series and in particular improves inference for shorter time series compared to local univariate models.
Such a model is expected to learn some advanced features (``embeddings'') exploiting information across time series. Once these advanced features are learned via $\Psi$, the global model is then used to forecast each time series \textit{independently}.
That is, although during training the model sees all the related time series together, the prediction  
is done by looking at each time series individually. 
Note that the embeddings learned in the global model are useful beyond the $N$ time series used in the training. 
This addresses the cold start problem in the sense that the global model can be used to provide forecasts for time series without historical values. Global models are also referred to as cross-learning  or panel models in econometrics and statistics and have been the subject of considerable study, e.g., via dynamic factor models~\citep{geweke1977dynamic}. 
	
\textbf{Multivariate model:} Here, a single model is learned for all $N$ time series using all available data, directly predicting the \textit{multivariate} target:
\begin{equation}
\label{eq:global_multivariate}
 p(\vZt{t+1:t+h} | \vZt{1:t},  \vXt{1:t+h}; \theta),\quad \theta = \Psi(\vZt{1:t},  \vXt{1:t+h}, \Phi).
\end{equation}
Note that the model also learns the dependency structure among the time series. Technically speaking, Eq.~\eqref{eq:global_multivariate} is a global multivariate model and a further distinction from local multivariate models, such as VARMA~\citep{Luetkepohl2005}, is possible. 

\paragraph{Remarks}
Note that in Eq.~\eqref{eq:general_set_up} and in the following model-specific cases we have chosen the multi-step ahead predictive distribution. 
We can always obtain a multi-step predictive distribution via a rolling one-step predictive distribution. In our discussion so far, we presented \textit{probabilistic forecast} models that learn the entire distribution of the future values. 
However, it may be desirable to model specific values such as the mean, median or some other quantile, instead of the whole probability distribution. These are called \textit{point-forecast} models and the optimal choice of the summary statistics to turn a probabilistic forecast into a point forecast depends on the metric used to judge the quality of the point forecast~\citep{KOLASSA2020208}.
More concretely, a point-forecast global univariate model learns a quantity $\hat{\mathbf{z}}_{i,t+1 :t+h}  = \Psi(\vzit{1:t}, \vXit{1:t+h},\Phi)$, where  $\hat{\mathbf{z}}_{i,t+1 :t+h}$ is some point estimate of the future values of the time series. Table~\ref{tab:category_sum} summarizes the various modelling option based on the forecast and model types.

\renewcommand{\arraystretch}{1.3}
\begin{table*}[t]
\begin{center}  
\small
\caption{Summary of \deep forecasting models based on forecast and model type. For one-step and multi-step forecasting models $h=1$ and $h>1$, respectively.}
\begin{tabularx}{0.83\textwidth}{c |c |c}
	\bottomrule
	Forecast type & Model type & Formulation\\
	\hline
	\multirow{3}{*}{Point}  & Local univariate & $\hat{\mathbf{z}}_{i,t+1 :t+h}  = \Psi(\vzit{1:t}, \vXit{1:t+h})$ \\\cline{2-3}
	& Global univariate&  $\hat{\mathbf{z}}_{i,t+1 :t+h}  = \Psi(\vzit{1:t}, \vXit{1:t+h},\Phi)$\\ \cline{2-3}
	& Multivariate &  $\hat{\mathbf{Z}}_{t+1 :t+h}   = \Psi(\vZt{1:t},  \vXt{1:t+h}, \Phi)$ \\ 
	\hline
	\multirow{3}{*}{Probabilistic}  & Local univariate & $P(\vzit{t+1:t+h} | \vzit{1:t}, \vXit{1:t+h} ; \theta_i),\quad \theta_i = \Psi(\vzit{1:t}, \vXit{1:t+h})$ \\\cline{2-3}
	& Global univariate &  $P(\vzit{t+1:t+h} | \vZt{1:t}, \vXt{1:t+h}; \theta_i),\quad \theta_i = \Psi(\vzit{1:t}, \vXit{1:t+h},\Phi)$ \\ \cline{2-3}
	& Multivariate &  $P(\vZt{t+1:t+h} | \vZt{1:t},  \vXt{1:t+h}; \theta),\quad \theta = \Psi(\vZt{1:t},  \vXt{1:t+h}, \Phi)$ \\ 
	\bottomrule
\end{tabularx}
\label{tab:category_sum}
\end{center}
\end{table*}
\renewcommand{\arraystretch}{1}

\subsection{Neural Network Architectures}
NNs are compositions of \textit {differentiable} 
functions formed from simple building blocks to learn an approximation of some unknown function from data. 
An NN is commonly represented as a directed acyclic graph consisting of nodes and edges. The edges between the nodes contain weights (also called parameters) that are learned from the data. The basic unit of every NN is a neuron (illustrated in Fig.~\ref{fig:node}), consisting of an input, an affine transformation with learnable weights and (optionally) a nonlinear activation function. Different types of NNs arrange these components in different ways. We refer to other reviews~\citep{Lim_2021} for more details on the main architectures. Here, we only offer a high-level summary for completeness, focusing instead on forecasting specific ingredients for NNs such as input processing and loss functions.

\subsubsection{Multilayer perceptron}
In multilayer perceptrons (MLPs) or synonymously feedforward NNs, layers of neurons are stacked on top of each other to learn more complex nonlinear representations of the data.
An MLP consists of an input and an output layer, while the intermediate layers are called \textit{hidden}.
The nodes in each layer of the network are fully connected to all the nodes in the previous layer.
The output of the last hidden layer can be seen as some nonlinear feature representation (also called an \emph{embedding}) obtained from the inputs of the network.
The output layer then learns a mapping from these nonlinear features to the actual target.
Learning with MLPs, and more generally with  NNs, can be thought of as the process of learning a nonlinear feature map of the inputs and the relationship between this feature map and the actual target. Figure \ref{fig:MLP} illustrates the structure of an MLP with two hidden layers. 
Modern incarnations of the MLP have added important details to alleviate problems like vanishing gradients~\citep{vanishing98}. For example, ResNet~\citep{he2015deep}, contains direct connections between hidden layers $\ell - 1$ and $\ell + 1$, skipping over the hidden layer $\ell$. 

One of the main limitations of MLPs is that they do not exploit the structure often present in the data in applications such as computer vision, natural language processing and forecasting.
Moreover, the number of inputs and outputs is fixed making them inapplicable to problems with varying input and output sizes as in forecasting. Next, we discuss more complex architectures that overcome these limitations, for which MLPs are often used as the basic building blocks.

\def\layersep{1.5cm}
\begin{figure}[t]
\centering

\begin{subfigure}[b]{0.44\textwidth}
\centering
\scalebox{0.8}{%
\begin{tikzpicture}[
    init/.style={ 
         draw, 
         circle, 
         inner sep=2pt,
         font=\Huge,
         join = by -latex
    },
    squa/.style={ 
        font=\Large,
        join = by -latex
    }
]
\begin{scope}[start chain=1]
    \node[on chain=1, label=above: \footnotesize Inputs, ] at (0,1.5cm)  (x1) {$\alpha_1$};
    \node[on chain=1, label=above: \footnotesize Weights, join=by o-latex] (w1) {$w_1$};
\end{scope}
\begin{scope}[start chain=2]
    \node[on chain=2] (x2) {$\alpha_2$};
    \node[on chain=2,join=by o-latex] {$w_2$};
    \node[on chain=2,init] (sigma) {$\displaystyle\Sigma$};
    \node[on chain=2,squa,label=above:{\parbox{2cm}{\centering \footnotesize Activation\\ function}}]   {$f(\cdot)$};
    \node[on chain=2,squa,label=above: \footnotesize Output,join=by -latex] {$\beta$};
\end{scope}
\begin{scope}[start chain=3]
    \node[on chain=3] at (0,-1.5cm) 
    (x3) {$\alpha_3$};
    \node[on chain=3, join=by o-latex]
    (w3) {$w_3$};
\end{scope}
\node[label=above:\parbox{2cm}{\centering \footnotesize Bias \\ $b$}] at (sigma|-w1) (b) {};
\draw[-latex] (w1) -- (sigma);
\draw[-latex] (w3) -- (sigma);
\draw[o-latex] (b) -- (sigma);
\end{tikzpicture}
}
\caption{Single node}
\label{fig:node}
\end{subfigure}
~
\begin{subfigure}[b]{0.53\textwidth}
\centering
\scalebox{0.95}{%
\begin{tikzpicture}[shorten >=1pt,->,draw=black!50, node distance=\layersep]
    \tikzstyle{every pin edge}=[<-,shorten <=1pt]
    \tikzstyle{neuron}=[circle,fill=black!25,minimum size=15pt,inner sep=0pt]
    \tikzstyle{input neuron}=[neuron, fill=green!50];
    \tikzstyle{output neuron}=[neuron, fill=red!50];
    \tikzstyle{hidden neuron}=[neuron, fill=blue!50];
    \tikzstyle{annot} = [text width=3em, text centered]

    \foreach \name / \y in {1,...,4}
        \node[input neuron] (I-\name) at (0,-\y) {$\alpha_{\y}$};

    \foreach \name / \y in {1,...,5}
        \path[yshift=0.5cm]
            node[hidden neuron] (H1-\name) at (\layersep,-\y cm) {};
            
    \foreach \name / \y in {1,...,3}
        \path[yshift=-0.5cm]
            node[hidden neuron] (Hk-\name) at (2*\layersep,-\y cm) {};

    \foreach \name / \y in {1,...,2}
     \path[yshift=-1cm]
        node[output neuron,pin={[pin edge={->}]right:$\beta_{\y}$}, right of=Hk-3] (O-\name) at (2*\layersep,-\y cm) {};

    \foreach \source in {1,...,4}
        \foreach \dest in {1,...,5}
            \path (I-\source) edge (H1-\dest);
            
    \foreach \source in {1,...,5}
        \foreach \dest in {1,...,3}
            \path (H1-\source) edge (Hk-\dest);

    \foreach \source in {1,...,3}
            \foreach \dest in {1,...,2}
	        \path (Hk-\source) edge (O-\dest);

    \node[annot,above of=H1-1, node distance=1cm] (hl) {\footnotesize Hidden layer 1};
    \node[annot,left of=hl] {\footnotesize Input layer};
    \node[annot,right of=hl] {\footnotesize Hidden layer 2};
    \node[annot,above of=O-1, node distance=2.5cm] (hl) {\footnotesize Output layer};
\end{tikzpicture}
}
\caption{MLP}
\label{fig:MLP}
\end{subfigure}
\caption{(a) Structure of a single node or neuron. An affine transformation is applied to the input followed by an activation function, i.e., $\beta = f\left(\sum \alpha_iw_i + b\right)$. The weights and bias parameters are learned during training. (b) Illustration of the MLP structure. Each circle in the hidden and output layers is a node, i.e., it applies an affine transformation followed by a nonlinear activation to the set of its inputs.\label{fig:SLPMLP}}
\end{figure}

\subsubsection{Convolutional neural networks}
Convolutional  neural networks (CNNs)~\citep{leCun1989conv} are a special class of  NNs that are designed for applications where inputs have a known ordinal structure such as images and time series~\citep{GBC2016}.
CNNs are \textit{locally} connected  NNs that use \textit{convolutional} layers to exploit the structure present in the input data 
by applying a convolution function to smaller neighborhoods of the input data.
Convolution here refers to the process of computing moving weighted sums by sliding the so-called \textit{filter} or \textit{kernel} over different parts of the input data. 
The size of the filter as well as how the filter is slid across the input are part of the hyperparameters of the model.
A nonlinear activation, typically ReLU \citep{glorot2011deep}, is then applied to the output of the convolution operation. 

In addition to convolutional layers, CNNs also typically use a \textit{pooling} layer to reduce the size of the feature representation as well as to make the features extracted from the convolutional layer more robust.
For example, a commonly used max-pooling layer, which is applied to the output of convolutional layers, extracts the maximum value of the features in a given neighborhood. 
Similarly to the convolution operation, the pooling operation is applied to smaller neighborhoods by sliding the corresponding filter over the input.
A pooling layer, however, does not have any learnable weights and hence both the convolution and the pooling layer are counted as one layer in CNNs.

Of particular importance for forecasting are the so-called \emph{causal} convolutions, defined as 
\[
h_j = \sum_{d \in \mathcal{D}} w_d \alpha_{j-d} \, ,
\]
where $h_j$ is the output of a hidden node, $\mathbf{\alpha}$ denotes the input, $\mathcal{D} = \{1,\ldots, n \}$ for some $n$, $|\mathcal{D}|$ is the \emph{width} of the causal convolution (or also called the receptive field) and $\mathbf{w}$ are the learnable 
parameters. In other words, causal convolutions are weighted moving averages which only take inputs into account which are before $j$ hence the reference to causality in its name. A variation are \emph{dilated} causal convolutions where we vary the index set $\mathcal{D}$, e.g., such that it does not necessarily contain consecutive values, but only every $k$-th value. Typically, 
these dilated causal convolutions are stacked on top of each other where the output of one layer of dilated causal convolutions is the input of another layer of causal convolution and the dilation grows by the depth of the NN. Figure \ref{fig:CNN} illustrates the general structure of a CNN with dilated causal convolutions.

\def\layersep{1.5cm}
\begin{figure}[t]
\centering

\begin{subfigure}[b]{0.48\textwidth}
\centering
\scalebox{0.7}{%
\begin{tikzpicture}[shorten >=1pt,->,draw=black!50, node distance=\layersep]
    \tikzstyle{every pin edge}=[<-,shorten <=1pt]
    \tikzstyle{blank}=[circle,minimum size=15pt,inner sep=0pt]
    \tikzstyle{neuron}=[circle,fill=black!25,minimum size=25pt,inner sep=0pt]
    \tikzstyle{input neuron}=[neuron, fill=green!50];
    \tikzstyle{output neuron}=[neuron, fill=red!50];
    \tikzstyle{hidden neuron}=[neuron, fill=blue!50];
    \tikzstyle{annot} = [text width=3em, text centered]

    \foreach \x in {1,...,5}
        \node[input neuron] (I-\x) at (1.5*\x, 0) {$\alpha_{\x}$};

   
    \foreach \name in {1,...,5}
            \node[hidden neuron] (H1-\name) at (1.5*\name cm, 1*\layersep) {}; 

    \foreach \name in {1,...,5}
            \node[hidden neuron] (H2-\name) at (1.5*\name cm, 2*\layersep) {}; 

    \foreach \name in {1,...,5}
            \node[output neuron] (O-\name) at (1.5*\name cm, 3*\layersep) {}; 

    \foreach \source in {3,4}
        \path (I-\source) edge (H1-4);

    \foreach \source in {1,2}
        \path (I-\source) edge (H1-2);

    \foreach \source in {2,4}
        \path (H1-\source) edge (H2-4);

    \foreach \source in {1,4}
        \path (H2-\source) edge (O-4);

\end{tikzpicture}
}
\caption{CNN} 
\label{fig:CNN}
\end{subfigure}
~
\begin{subfigure}[b]{0.48\textwidth}
\centering
\scalebox{0.7}{%
\begin{tikzpicture}[shorten >=1pt,->,draw=black!50, node distance=\layersep]
    \tikzstyle{every pin edge}=[<-,shorten <=1pt]
    \tikzstyle{blank}=[circle,minimum size=15pt,inner sep=0pt]
    \tikzstyle{neuron}=[circle,fill=black!25,minimum size=25pt,inner sep=0pt]
    \tikzstyle{input neuron}=[neuron, fill=green!50];
    \tikzstyle{output neuron}=[neuron, fill=red!50];
    \tikzstyle{hidden neuron}=[neuron, fill=blue!50];
    \tikzstyle{annot} = [text width=3em, text centered]

    \foreach \name / \y in {1/1, 2/2, 3/3}
        \node[input neuron] (I-\name) at (1.5*\name, 0) {$\alpha_{\y}$};

    \node[hidden neuron] (H1-0) at (0 cm, \layersep) {$\mathbf{h}_0$};	
    \foreach \name / \y in {1/1, 2/2, 3/3}
            \node[hidden neuron] (H1-\name) at (1.5*\name cm, \layersep) {$\mathbf{h}_{\y}$};
    \node[hidden neuron] (H1-4) at (1.5*4 cm, \layersep) {$\mathbf{h}_4$};	
    
    \foreach \name / \y in {1/1, 2/2, 3/3}
            \node[blank] (O-\name) at (1.5*\name cm, 2*\layersep) {$\beta_{\y}$};

    \foreach \source in {1,...,3}
        \path (H1-\source) edge (O-\source);
        
    \foreach \source in {1,...,3}
        \path (I-\source) edge (H1-\source);
            
    \foreach \source/\dest in {0/1, 1/2, 2/3, 3/4}
        \path (H1-\source) edge (H1-\dest);

\end{tikzpicture}
}
\vspace{4mm}
\caption{RNN}
\label{fig:RNN}
\end{subfigure}
\caption{(a) Structure of a CNN consisting of a stack of three causal convolution layer. The input layer (green) is non-dilated and the other two are dilated. (b) Structure of an unrolled RNN. At each timestep $t$ the network receives an external input $\alpha_t$ and the output of the hidden units from the previous time step $\mathbf{h}_{t-1}$. The hidden units all share the same weights. The internal state of the network is updated to $\mathbf{h}_{t}$ that is going to play the role of the previous state in the next timestep $t+1$. Finally, the network outputs $\beta_t$ which is a function of $\alpha_t$ and $\mathbf{h}_{t}$.}
\label{fig:cnn_rnn}

\end{figure}

\subsubsection{Recurrent neural networks}
Recurrent neural networks (RNNs) are NNs specifically designed to handle sequential data that arise in applications such as time series, natural language processing and speech recognition. 
The core idea consists of connecting \textit{recurrently} the NNs' hidden units back to themselves with a time delay~\citep{jordan1986serial, jordan1989serial}. 
Since hidden units learn some kind of feature representations of the raw input, feeding them back to themselves can be interpreted as providing the network with a dynamic memory.
One crucial detail here is that the same network is used for all timesteps, i.e., the weights of the network are shared across timesteps. 
This weight-sharing idea is similar to that in CNNs where the same filter is used across different parts of the input.
This allows the RNNs to handle sequences of varying length during training and, more importantly, generalize to sequence lengths not seen during training.
Figure \ref{fig:RNN} illustrates the general structure of an (unrolled) RNN.

Although RNNs have been widely used in practice, training them is difficult given that they are typically applied to long sequences of data. A common issue while training very deep NNs by gradient-based methods using back-propagation is that of vanishing or exploding gradients ~\citep{Pascanu2013} which renders learning challenging. ~\citet{hochreiter1997long} proposed 
Long short-term memory networks (LSTM) to address this problem. Similar to Resnet, via the skip-connections, LSTMs (and a simplified version Gated recurrent units (GRU)~\cite{Cho2014GRU}) always offer a path where the gradient does not vanish or explode.

\subsubsection{Transformer}
A more recent architecture is based on the \textit{attention} mechanism which has received increased interest in other sequence learning tasks~\citep{chorowski2014end, chorowski2015attention, vaswani2017attention, li2019enhancing} for its ability to improve on long sequence prediction tasks over RNNs.
One natural way to address this issue is to learn more than one feature representation (contrary to RNNs), e.g., one for each time step of the input sequence and decide which of these representations are useful to predict the current element of the target sequence. 
\citet{BahdanauCB14} suggest using a weighted sum of the representations where the weights are jointly learned along with the feature representation learning and the prediction. 
Note that at each time step in the prediction, one needs to learn a separate set of weights for the representations.
This is essentially training the predictor to learn to which parts of the input sequence it should pay \textit{attention} to produce a prediction. 
This attention mechanism has been shown to be instrumental for the state of the art in speech recognition and machine translations tasks~\citep{chorowski2014end, chorowski2015attention}.
Inspired by the success of attention models, ~\citet{vaswani2017attention} developed the so-called \textit{Transformer} model and showed that attention alone is sufficient, 
thus making the training amenable for parallelization and large number of parameters~\citep{gpt3,Devlin2018}. In the literature, the term Transformer can refer to both the specific model and to the overall architecture as well.

\subsection{Input Transformations}\label{sec:input_transformation}
The careful handling of the input (parameters $\alpha_t$ in Fig.~\ref{fig:SLPMLP} and \ref{fig:cnn_rnn}) is a practically important ingredient for deep learning models in general and \deep forecasting models in particular. \Deep forecasting models are most commonly deployed as so-called global models (see Section~\ref{ssec:formalization}), which means that the weights of the NN are trained across the panel of time series. Hence, it is important that the scale of the input is comparable. Standard techniques such as mean-variance scaling carry over to the forecasting setting. In practice, it is important to avoid leakage of future values in normalization schemes, so that mean and variance are taken over past windows (similar to causal convolutions).  

Traditionally, the forecasting literature has used transformations such as the Box-Cox, i.e., 
\begin{equation}\label{eq:boxcox}
h = \frac{z^{\lambda}-1}{\lambda},
\end{equation}
where $z$ is the input of the transformation, $h$ is the output and $\lambda$ is a free parameter. Box-Cox is a popular heuristic to have the input data more closely resemble the Gaussian distribution. A Box-Cox transformation can be readily integrated into an NN, with the free parameter $\lambda$ optimized as part of the training process jointly with the other 
parameters of the network. More sophisticated approaches based on probability integral transformation (PIT) or Copulas are similarly possible, see e.g.,~\citep{janke2021implicit} (and references therein) for a recent example.

A further standard technique is 
the discretization of input into categorical values or \emph{bins}, for example by choosing the number and borders of bins such that 
each bins contains equal mass, see e.g.,~\citep{rabanser2020effectiveness} for an example in forecasting.

We note that any input transformation must be reversed also to obtain values in the actual domain of interest. It is a choice for the modeller where/when to apply this reversal. Two extreme choices are to have transformation of the input and output fully outside the NN or have the input transformations as part of the NN and hence be subjected to learning.

\subsection{Output Models and Loss Functions}\label{sec:output_transformation}

Similar to the input, the output ($\beta_t$ in Fig.~\ref{fig:SLPMLP} and~\ref{fig:cnn_rnn}) deserve a special discussion. Closely related  is the question on the choice of loss function which we use to train a NN.
The simplest form of an output is a single value, also referred to as a \emph{point} forecast. 
For this case, the output $\hat{z}_{i,t}$ is the best (w.r.t.\ the chosen loss function) estimate for the true value $z_{i,t}$. Standard regression loss functions (like $\ell_p$ losses with their regularized modifications) can be used or more sophistication accuracy metrics specifically geared towards forecasting such as the MASE, sMAPE or others~\citep{Hyndman06anotherlook}.

\begin{figure}
\includegraphics[width=\textwidth]{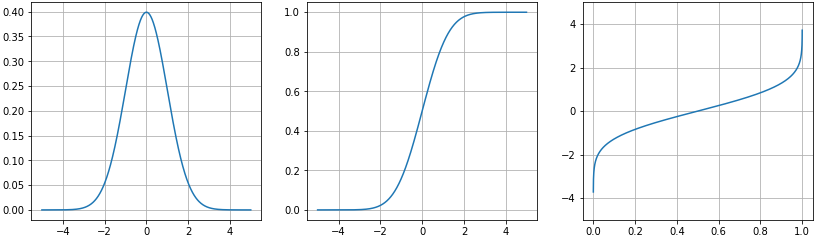}
\caption{For a Gaussian distribution, its density function $f$ is on the left-hand panel, the corresponding cumulative density function $F$ (the primitive integral of $f$) in the central panel and the quantile function $F^{-1}$ on the right-hand panel. 
\label{fig:pdf_cdf_quantile}}
\end{figure}

As remarked in Section~\ref{ssec:formalization}, a point estimate $\hat{z}_{i,t}$ can be seen as a particular realization from a probabilistic estimate of $p(\zit{t})$. Depending on the accuracy metric used in forecasting, a different realization may be appropriate~\citep{KOLASSA2020208}. So, even for obtaining point forecasts, \emph{probabilistic} forecasts are important. More importantly, forecasts are often used in downstream optimization problem where some form of \emph{expected} cost is to be minimized and for this, an estimate of the entire probability distribution is required. The probability distribution can be represented equivalently by its probability density function (PDF), the cumulative density function (CDF) or its inverse, the quantile function. 
Fig.~\ref{fig:pdf_cdf_quantile} contains a visualization of the different representations for the Gaussian distribution. Across the \deep forecasting landscape, most approaches (e.g.,~\citep{salinas2020deepar,gasthaus2019probabilistic,rasul2020multi,rangapur21,rangapuram2018deep}), have chosen the PDF and quantile function to represent $p(\zit{t})$ and we will discuss general 
recipes next. Since the CDF has typically not been chosen to represent $p(\zit{t})$, we do not discuss it further.

\subsubsection{PDF}
Arguably the most common way to represent a probability distribution in forecasting is via its PDF. The literature contains 
examples of using the standard parametric distribution families to represent probabilistic forecasts. For example, the output layer of an NN may produce the mean and variance parameter of a Gaussian distribution. So, the parameter $\beta_t$ in Fig.~\ref{fig:SLPMLP} and~\ref{fig:cnn_rnn} is a two-dimensional vector corresponding to $\mu_t$ and $\sigma_t$ of a Gaussian distribution. We typically achieve $\sigma_t \geq 0$ by mapping the corresponding parameter through a softplus function. For the loss function, a natural choice is the negative log-likelihood (NLL) since a PDF allows to readily compute the likelihood of a point under it. 

Beyond Gaussian likelihood, a number of differentiable parametric distributions have been used in the literature depending on the nature of the forecasting problem, e.g., the student-t distribution or the Tweedie distribution for continuous data, the negative binomial distribution for count data and more flexible approaches via mixtures of Gaussian. Although forecasting is most commonly done for domains of numerical values (i.e., we assume $z_{i,t}$ to be in $\mathbb{R}$ or $\mathbb{N}$), other distributions such as the multinomial have also been employed successfully in forecasting even though they have no notion of the order on the domain  \citep{rabanser2020effectiveness}. The deployment of a multinomial distribution requires a binning of the input values (see Section~\ref{sec:input_transformation}). An alternative approach is to cut the output space in bins and treat each of them as a uniform distribution, while modelling the tails with a parametric distribution \citep{ehrlich2021spliced}, this results in a piecewise linear CDF.

\subsubsection{Quantile function}
Another representation of $p(\zit{t})$ is via the quantile function which has a particular importance for forecasting. Often, a particular quantile is of practical interest. For example, in a simplified supply chain scenario for inventory control, there is a direct correspondence between the chosen quantile and a safety stock level in the newsvendor problem~\citep{petropoulos2020forecasting}. 

So naturally, estimating the quantiles directly via quantile regression approaches~\citep{koenker2005quantile} is a common choice in forecasting either via choosing a single quantile (in a point-forecasting approach) or multiple quantiles simultaneously~\citep{eisenach2020mqtransformer,wen2017multi}. Essentially, this discretizes the quantile function and estimates specific points only. A common choice for the loss function is the quantile loss or pinball loss. For the $q$-th quantile and $F^{-1}$ the quantile function, the quantile loss is defined as 
\begin{equation}\label{eq:QL}
	\mathrm{QS}_q\left(\hat{F}_{i,t}^{-1}(q), z_{i,t}\right) \defeq 2\left(\mathbbm{1}_{\{z_{i, t} \leq \hat{F}_{i,t}^{-1}(q)\}}-q\right)\left(\hat{F}_{i,t}^{-1}(q) - z_{i,t}\right),
\end{equation}
where $\mathbbm{1}_{\{\text{\textit{cond}}\}}$ is the indicator function that is equal to 1 if cond is true and 0 otherwise.
The output of the NN is $\hat{F}_{i,t}^{-1}(q)$, i.e., the estimated value of the $q$-th quantile. For $q=0.5$ this reduces to the median of the forecast distribution and is a common choice of point forecasts.

As an alternative to a quantile regression approach, we can make a parametric assumption on the quantile function and estimate it directly. The main requirements for modelling a quantile function are that its domain should be constrained to $[0,1]$ and the function should be monotonically increasing. This can be achieved easily via linear splines for example, so the output of the NN's last layers are the corresponding free parameters. 
For the loss function, a rich theory around the continuous ranked probability score (CRPS) exists~\citep{matheson76,gneiting2007probabilistic} and CRPS can be used 
as a loss function directly. CRPS can be defined~\citep{laio07} to summarize all possible quantile losses as

\begin{align}\label{defn:crps}
	\mathrm{CRPS}(\hat{F}_{i,t}, z_{i,t}) &\defeq
	\int_0^1\mathrm{QS}_q\left(\hat{F}_{i,t}^{-1}(q), z_{i,t}\right) \ dq.	
\end{align}
Multivariate extensions such as the energy score~\citep{gneiting2007probabilistic} exist.

\begin{figure}
\includegraphics[width=.3\textwidth]{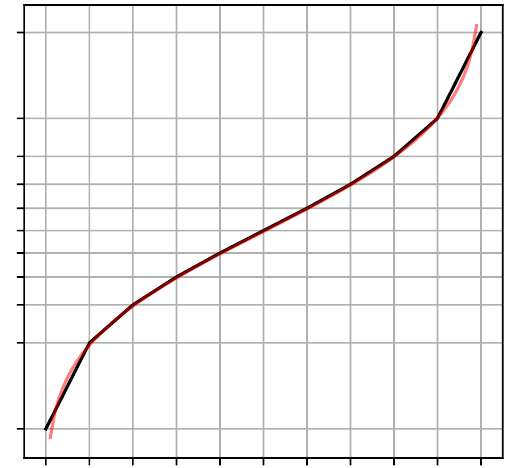}\hfill
\includegraphics[width=.3\textwidth]{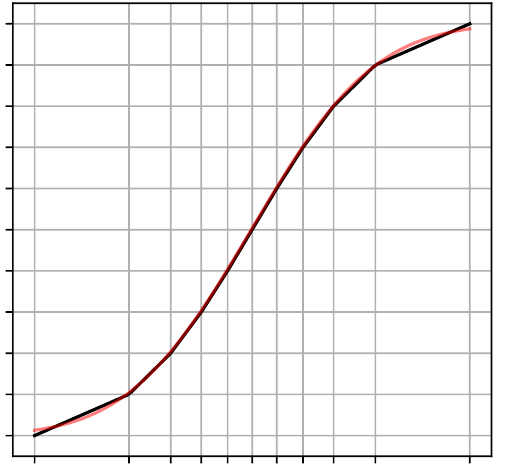}\hfill
\includegraphics[width=.3\textwidth]{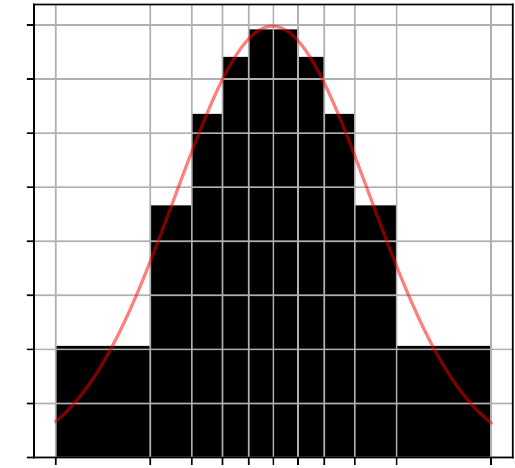}
\caption{An illustration how a quantile function parametrized by linear splines (left panel) corresponds to a piece-wise linear CDF (middle) which in turn corresponds to a piece-wise constant PDF as assumed in an adaptive binning strategy (right panel).\label{fig:splines}}
\end{figure}

Interestingly, a popular discretization strategy, adaptive binning, used 
with 
multinomial distributions corresponds to quantile functions parametrized by piece-wise linear splines, see Fig.~\ref{fig:splines}.

\subsubsection{Further approaches}
\label{ssec:further}

The recent deep learning literature contains more advanced examples for density estimation, most prominently via Generalized Adversarial Networks (GANs). We discuss them in Section~\ref{sec:gans} and discuss \emph{normalizing flows} here which have arguably resonated more strongly in forecasting. Normalizing flows are invertible NNs that transform a simple distribution to a more complex output distribution. Invertibility guarantees the conservation of probability mass and allows the evaluation of the associated density function everywhere. 
The key observation is that the probability density of an observation $\zit{t}$ can be computed using the change of variables formula:
\begin{equation}
	\label{eq:density}
	p(\zit{t}) =  p_{y_{i,t}}(f^{-1}(\zit{t}) )| \text{det}[\text{Jac}{f_{i,t}^{-1}}{\zit{t}}]|,
\end{equation} 
where the first term $p_{y_{i,t}}(f^{-1}(\zit{t}) )$ is the (in general simple) density of a variable $y_{i,t}$, and the second is the absolute value of the determinant of the Jacobian of  $f_{i,t}^{-1}$,  evaluated at $\zit{t}$. 

The invertible function $f$ is typically parametrized by an NN. A particular instantiation is the Box-Cox transformation, Eq.~\eqref{eq:boxcox}. The field of normalizing flows (e.g.,~\citep{realnvp,glow,TAN}) studies invertible NNs that typically transform isotropic Gaussians to more complex data distributions. The choice of a particular instantiation of $f$ can facilitate the computation of the likelihood of a given point when the NLL is amenable as a loss function. Alternatively, generating samples may be computationally more viable for other instantiations (this is typically the cases with generative adversarial networks as well). In this case, the NLL can be replaced by other loss functions such as CRPS.

A number of 
extensions are possible. For example, more complex models for $p(\zit{t})$ are possible such as 
Hidden Markov Models or Linear Dynamical Systems. NNs can output the free parameters of these models but then need to be combined with the learning and inference schemes associated with these models, such as Kalman Filtering/Smoothing in the case of Linear Dynamical System~\citep{rangapuram2018deep,de2020normalizing} or the Forward/Backward Algorithm in the case of Hidden Markov Models~\citep{redsds}. Another avenue is to relax constraints on the representation of $p(\zit{t})$ to obtain closely related objects with more favorable computational properties. For example, \emph{energy based models} (EBMs) approximate the unnormalized log-probability~\citep{LeCun06atutorial,hinton02}. EBMs 
perform well in learning high dimensional
distributions at the cost of being difficult to train~\citep{song2021train} and have been employed in forecasting~\citep{rasul2021autoregressive}.

\subsection{Archetypical Architectures}
\label{sec:fct_models}

\begin{figure}
\centering
\includegraphics[width=0.32\textwidth]{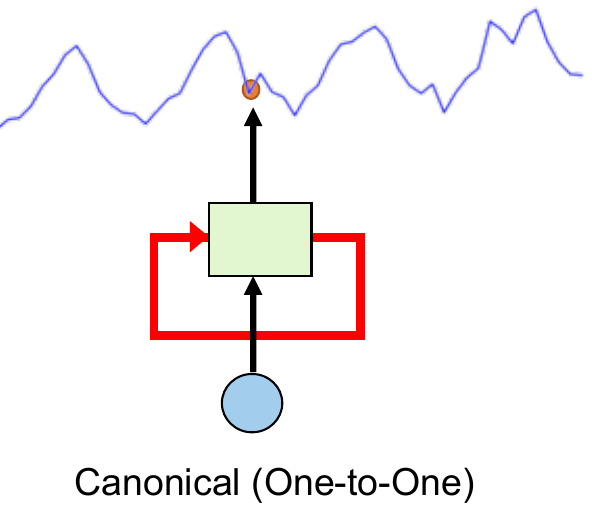}\hspace{1cm}
\includegraphics[width=0.32\textwidth]{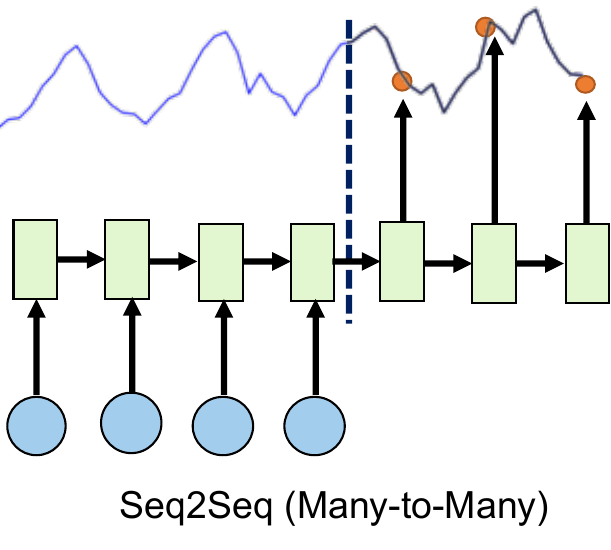}
\caption{Canonical versus sequence-to-sequence models.\label{fig:gen_disc}}
\end{figure}

With all key components in place, in this section we present in more details popular forecasting architectures. 
In particular we focus on the widely-used RNN-based architecture that takes as input its previous hidden state, the currently available information and produces an one-step ahead estimate of the target time series. There are 
subtle details on how to handle a multi-step unrolled model during training (e.g.,~\citep{lamb16}), which we will skip over. We further examine the sequence-to-sequence (seq2seq) modelling approach where the model takes an \emph{encoding sequence} as input and maps it to a \emph{decoding} sequence (of predetermined length) on which the loss is computed against the actual values $\mathbf{z}$ during training. \def\layersep{1.5cm}
\begin{wrapfigure}[12]{r}{0.5\textwidth}
\vspace{-9mm}
\centering
\scalebox{0.85}{%
\begin{tikzpicture}[shorten >=1pt,->,draw=black!50, node distance=\layersep]
    \tikzstyle{every pin edge}=[<-,shorten <=1pt]
    \tikzstyle{blank}=[circle,minimum size=1.5cm,inner sep=0pt]
    \tikzstyle{neuron}=[circle,fill=black!25,minimum size=25pt,inner sep=0pt]
    \tikzstyle{input neuron}=[neuron, fill=green!50];
    \tikzstyle{output neuron}=[neuron, fill=red!50];
    \tikzstyle{hidden neuron}=[neuron, fill=blue!50];
    \tikzstyle{annot} = [text width=3em, text centered]

    \foreach \name / \y in {1/{t-1}, 2/{t}, 3/{t+1}}
        \node[input neuron] (I-\name) at (1.7*\name, 0) {$\alpha_{\y}$};

    \node[hidden neuron] (H1-0) at (0 cm, \layersep) {$\mathbf{h}_{t-2}$};       
    \foreach \name / \y in {1/{t-1}, 2/{t}, 3/{t+1}}
            \node[hidden neuron] (H1-\name) at (1.7*\name cm, \layersep) {$\mathbf{h}_{\y}$};
    \node[hidden neuron] (H1-4) at (1.7*4 cm, \layersep) {$\mathbf{h}_{t+2}$};   
    
    \foreach \name / \y in {1/{t}, 2/{t+1}, 3/{t+2}}
            \node[blank] (O-\name) at (1.7*\name cm, 2.5*\layersep) {$(\hat{\mu}_{\y}, \hat{\sigma}_{\y})$};

    \foreach \source in {1,...,3}
        \path (H1-\source) edge (O-\source);

    \foreach \source/\dest in {1/2, 2/3, 3/4}
        \path (O-\source) edge[dotted] (H1-\dest);

    \foreach \source in {1,...,3}
        \path (I-\source) edge (H1-\source);

    \path (I-1) edge (H1-3);
    \path (I-2) edge (H1-4);
            
    \foreach \source/\dest in {0/1, 1/2, 2/3, 3/4}
        \path (H1-\source) edge (H1-\dest);

\end{tikzpicture}
}
\caption{DeepAR: The model outputs parameters of a previously chosen family of distributions. Samples from this distribution can be fed back into the model during prediction (dotted lines) or in case of $\alpha_t$ being missing. 
}
\label{fig:DeepAR}
\end{wrapfigure}
A typical instance in the training set in this approach consists of the target and covariate values up to a certain point in time $t$ as the encoding sequence and the outputs of the NN are a predetermined number of target values after time $t$. 
Figure~\ref{fig:gen_disc} contrasts both approaches. In  the following we present two popular \deep forecasting models, DeepAR and MQRNN/MQCNN, in some details to illustrate the core concepts. They represent the one-step-ahead RNN-based and seq2seq approach, respectively.

\subsubsection{DeepAR}\label{sec:deepar}
Among the first of the modern \deep forecasting models is DeepAR~\citep{salinas2020deepar}, a global univariate model (see Table~\ref{tab:category_sum}) that consists of an RNN backbone (typically an LSTM).\footnote{\citet{hewamalage2019recurrent} provide an overview specifically targeted at RNNs for forecasting.} 
The input of the model is a combination of lagged target values and relevant covariates. 
The output is either a point forecast with a standard loss function or, in the basic variant, a probabilistic forecast via the parameters of a PDF (e.g., $\mu$ and $\sigma$ of a Gaussian distribution), where the loss function is then the NLL. The output modelling of DeepAR has been the subject of follow-up work, e.g.,~\citet{JEON2021} propose a Tweedie loss, \citet{mukherjee2018armdn} propose a mixture of Gaussians as the distribution and domain specific feature processing blocks. Figure~\ref{fig:DeepAR} summarizes the architecture. The dotted arrows in the picture correspond to drawing a sample that can be used as alternative input (as a lagged target) during training (even though $\alpha_t$ may be available or in the case where an $\alpha_t$ is missing) and during prediction to obtain multi-step ahead forecasts. 

It is also possible to change the output of DeepAR to model the quantile function and use CRPS, Eq.~\eqref{defn:crps}, as the loss function~\citep{gasthaus2019probabilistic,kan22, gouttes2021probabilistic}. While this in general computationally challenging, special cases are amendable for practical computation. For example,  we can assume a parametrization of the quantile function by linear isotonic regression splines:
\begin{align}\label{eq:linear_spline}
s(q; \gamma, b, d) = \gamma + \sum_{\ell = 0}^{L}b_\ell (q - d_\ell)_+ \;
\end{align}
where $q\in [0,1]$  is the quantile level, $\gamma \in \mathbb{R}$ is the intercept term, $b \in \mathbb{R}^{L+1}$ are weights describing the slopes of the function pieces, $d \in \mathbb{R}^{L+1}$ is a vector of knot positions, $L$ the number of pieces of the spline and $(x)_+ = \max(x,0)$ is the ReLU function. In order for $s(\cdot)$ to represent a quantile function we need to guarantee its monotonicity and restrict its domain to $[0,1]$. Both of these constraints can readily be achieved using standard NN tooling using a reparametrization of Eq.~\eqref{eq:linear_spline}, while CRPS can be solved in closed form for linear splines (see \citep{gasthaus2019probabilistic}).

\paragraph{Bag of tricks.} While the general setup of DeepAR is straightforward, a number of algorithmic optimizations turn it into a robust, general-purpose forecasting model. The handling of missing values via sample replacement from the probability distribution is one such example. Another one is oversampling of ``important'' training examples during training, where importance typically corresponds to time series with larger absolute values. Adding lagged values further help improve predictive accuracy. Lags can be 
chosen heuristically based on the frequency of the time series. For example, in a time series with daily frequency, a lag of 7 days often helps. Similarly, covariates corresponding to calendar events (e.g., indicator variables for weekends or holidays) can help further.

\subsubsection{MQRNN/MQCNN}\label{sec:mqcnn}
As an example for another type of \deep forecasting model, we discuss the multi-horizon quantile recurrent forecaster (MQRNN)~\citep{wen2017multi}  next which was conceived concurrently to DeepAR. Contrary to DeepAR, it is most naturally deployed 
as a discriminative, seq2seq model in a quantile regression setting. For each time point $t$ in the forecast horizon, MQRNN outputs a chosen number 
of estimates for corresponding quantiles and the loss function in MQRNN is Eq.~\eqref{eq:QL}, i.e., the pinball loss summed over all quantiles and time points.

While MQRNN can use multiple configurations, often a CNN-based architecture is chosen in practice in the encoder (MQCNN) for computational efficiency reasons over RNN-based methods and two MLPs in the decoder. The first MLP captures all inputs during the forecast horizon and the context provided by the encoder. A second, local MLP applies only to specific horizons for which it uses the corresponding available input and the output of the MLP. A further innovation provided by MQCNN is the training scheme via the so-called \emph{forking sequences} where the model forecasts
 by placing a series of decoders with shared parameters at each timestep in the encoder. Thus, the model can structurally 
forecast at each timestep, while the optimization process is stabilized by updating the gradients from the sequences together. An additional component of MQRNN is a local MLP component that aims to model spikes and events specifically. 
 

\section{Literature review}
\label{sec:lit_review}
In the prior section, we provided an in-depth introduction to selected, basic topics. Building on these topics, we survey the literature on modern \deep forecasting models more broadly in this section. Given the breadth of the literature available, our selection is necessarily subjective.

We proceed as follows. In Section~\ref{ssec:prob_models} we present probabilistic forecasting models, both one-step and multi-step. Similarly, in Section~\ref{ssec:point_models} we summarize point forecast models. We remark that, after Section~\ref{sec:building_blocks}, we have recipes at hand to turn an one-step ahead forecasting model into a multi-step forecasting model and a point forecasting model into a probabilistic model. We discuss hybrids of deep learning with state space models in Section~\ref{sec:state_space}, multivariate forecasting in Section~\ref{sec:multivariate}, physics-based model in Section~\ref{sec:physics}, \gl models in Section~\ref{sec:global_local}, models for intermittent time series in Section~\ref{sec:intermittent} and generative adversarial networks for forecasting in Section~\ref{sec:gans}. We close this section with an overview of the large number of available models in Section~\ref{ssec:summary_models} where we also provide guidelines on where to start the journey with \deep forecasting models.

\subsection{Probabilistic Forecast Models}
\label{ssec:prob_models}

\subsubsection{One-step forecast}\label{sec:auto_regressive}

The DeepAR model presented in Sec.~\ref{sec:deepar}, is an example of one-step canonical forecasting model. In its base variant, DeepAR is a global univariate model which learns a univariate distribution; we discuss multivariate extensions in Sec.~\ref{sec:multivariate}. DeepAR can be equipped with outputs representing a parametrized PDF including Gaussian Mixture Distributions~\citet{mukherjee2018armdn} or quantile functions~\citet{gasthaus2019probabilistic}.

\citet{rasul2021autoregressive} propose TimeGrad which, like DeepAR, is an RNN model using LSTM or GRU cells for which samples are drawn from the data distribution at each time step, with the difference that in TimeGrad the RNN conditions a diffusion probabilistic model \citep{sohl2015deep} which allows the model to easily scale to multivariate time series and accurately use the dependencies between dimensions.
Replacing the RNN-backbone of DeepAR with dilated causal convolutions has been proposed as both point and probabilistic forecasting models~\citep{bischof19,alexandrov2019gluonts,van2016wavenet}.

\subsubsection{Multi-step forecast}



Contrary to the some of the models in Section~\ref{sec:auto_regressive} which produce one-step ahead forecasts, multi-step forecasts can be obtained directly with a seq2seq architecture. In Section~\ref{sec:mqcnn}, we reviewed the MQRNN/MQCNN architecture~\citep{wen2017multi} as a seq2seq architecture for probabilistic forecasting. The main advantage of seq2seq over one-step ahead forecast models is that the decoder architecture can be chosen to output all future target values at once. This removes the need to unroll over the forecast horizon which can lead to \emph{error accumulation} since early forecast errors propagate through the forecast horizon. Thus, the decoder of seq2seq forecasting models is typically an MLP while other architectures are also used for the encoder~\citep{wen2017multi, oreshkin2019n}.

~\citet{wen2019deep} extended the MQCNN model with a generative quantile copula. This model learns the conditional quantile function that maps the quantile index, which is a uniform random variable conditioned on the covariates, to the target. During training, the model draws the quantile index from a uniform distribution. This turns MQCNN into a generative, marginal quantile model. The authors combine this approach with a Gaussian copula to draw correlated marginal quantile index random values. They show that the Gaussian copula component improves the forecast at the distribution tails. \citet{chen2019probabilistic} proposed DeepTCN, another seq2seq model where the encoder is the dilated causal convolution with residual blocks, and the decoder is simply an MLP with residual connections. Structure-wise, DeepTCN is almost the same as the basic structure of MQCNN~\citep{wen2017multi}, i.e., without the local MLP component that aims to model spikes and events. 

\citet{park2021learning} propose the incremental quantile functions (IQF), a flexible and efficient distribution-free quantile estimation
framework that resolves quantile crossing with
a simple NN layer. A seq2seq encoder-decoder structure is used although the method can be readily applied to recurrent models with one-step ahead forecasts \cite{salinas2020deepar}.  IQF is trained using the CRPS loss (Eq.~\eqref{defn:crps}) similar to \cite{gasthaus2019probabilistic}. 

A combination of recurrent and encoder-decoder structures has also been explored. In \citep{zhu2017deep}, the authors use an LSTM with Monte Carlo dropout as both the encoder and decoder. However, unlike other models that directly use RNNs to generate forecasts, the learned embedding at the end of the decoding step is fed into an MLP prediction network and is combined with other external features to generate the forecast. Along a similar line, \citet{laptev2017} employ an LSTM as a feature extractor (LSTM autoencoder), and use the extracted features, combined with external inputs to generate the forecasts with another LSTM.

\citet{van2016wavenet} introduced the WaveNet architecture, a generative model
for speech synthesis, which uses dilated causal convolutions to
learn the long range dependencies important for audio signals. Since this
architecture is based on convolutions, training is very efficient on GPUs –
prediction is still sequential and further changes are necessary for fast inference. Adaptations of WaveNet for forecasting are available \cite{alexandrov2019gluonts}.

\subsection{Point Forecast Models}
\label{ssec:point_models}

Point forecast models do not model the probability distribution of the future values of a time series but rather output directly a point forecast that typically corresponds to a summary statistic of the predictive distribution. We have discussed generic recipes on how to turn a point forecasting model into a probabilistic forecasting model in Section~\ref{sec:output_transformation} and the literature contains further examples (see e.g.,~\citep{hasson21,schaar21} for recent complementary approaches). 

\subsubsection{One-step forecast}
A considerable amount of attention of the community is dedicated to one-step forecasting. LSTNet \citep{lai2018modeling} is a model using a combination of a CNN and an RNN. Targeting multivariate time series, LSTNet uses a convolution network (without pooling) to extract short-term temporal patterns as well as correlations among variables. The output of the convolution network is fed into a recurrent layer and a temporal attention layer which, combined with the autoregressive component, generates the final point forecast. While LSTNet uses a standard point forecast loss function, it can readily be turned into a probabilistic forecast model using the components described in Sec.~\ref{sec:building_blocks}, e.g., by modifying LSTNet to output the parameters of a probability distribution and using NLL as a loss function. \citet{qiu2014ensemble} proposed an ensemble of deep belief networks for forecasting. The outputs of all the networks is concatenated and fed into a support vector regression model (SVR) that gives the final prediction. The NNs and the SVR are not trained jointly though. \citet{hsu2017time} proposed an augmented LSTM model which combines autoencoders with LSTM cells. The input observations are first encoded to latent variables, which is equivalent to feature extraction, and are fed into the LSTM cells. The decoder is an MLP which maps the LSTM output into the predicted values. 
For point forecast multivariate forecasting, \citet{yoo2021attention} proposed time-invariant attention to learn the dependencies between the dimensions of the time series and use them with a convolution architecture to model the time series.

Building upon the success of CNNs in other application domains, \citet{borovykh2017conditional} proposed an adjustment to WaveNet~\citep{van2016wavenet} that makes it applicable to conditional forecasting. They evaluated their model on various datasets with mixed results, concluding that it can serve as a strong baseline and that various improvements could be made. In a similar vein, inspired by the Transformer architecture~\citep{vaswani2017attention} 
\citet{song2018attend} proposed an adjustment that makes the architecture applicable to time series. Their method is applied to both regression and classification tasks.

\subsubsection{Multi-step forecast}
N-BEATS~\citep{oreshkin2019n} is an NN architecture purpose-built for the forecasting task that relies on a deep, residual stack of MLP layers to obtain point forecasts. The basic building block in this architecture is a forked MLP stack that takes the block input and feeds the intermediate representation into separate MLPs to learn the parameters of the context (the authors call it backcast) and forecast time series models. The residual architecture removes the part of the context signal it can explain well before passing to the next block and adds up the forecasts. The learned time series model can have free parameters or be constrained to follow a particular, functional form. Constraining the model to trend and seasonality functional forms does not have a big impact on the error and generates models whose stacks are interpretable since the trend and seasonality components of the model can be separated and analyzed.
N-BEATS has also been interpreted as a meta-learning model \citep{oreshkin2020meta}, where the repeated application of residual blocks can be seen as an inner optimization loop. N-BEATS generalizes better than other architectures when trained on a source dataset (e.g.,\ M4-monthly) and applied to a different target datasets (e.g.,\ M3-monthly).

\citet{lv2014traffic} propose a \emph{stacked autoencoder} (SAE) architecture to learn features from spatio-temporal traffic flow data. On top of the autoencoder, a logistic regression layer is used to output predictions of the traffic flow at all locations in a future time window. The resulting architecture is trained layer-wise in a greedy manner. 
The experimental results show that the method significantly improves over other shallow architectures, suggesting that the SAE is capable of extracting latent features regarding the spatio-temporal correlations of the data. In the same context of spatio-temporal forecasting and under the seq2seq framework, \citet{li2018diffusion} proposed the Diffusion Convolutional Recurrent NN (DCRNN). Diffusion convolution is employed to capture the dependencies on the spatial domain, while an RNN is utilized to model the temporal dependencies. Finally, \citet{asadi2019spatial} proposed a framework where the time series are decomposed in an initial preprocessing step to separately feed short-term, long-term, and spatial patterns into different components of a NN. Neighbouring time series are clustered based on their similarity of the residuals as there can be meaningful short-term patterns for spatial time series. Then, in a CNN based architecture, each kernel of a multi-kernel convolution layer is applied to a cluster of time series to extract short-term features in neighbouring areas. The output of the convolution layer is concatenated by trends and is followed by a convolution-LSTM layer to capture long-term patterns in larger regional areas.

\citet{bandara2017forecasting} addressed the problem of
predicting a set of disparate time series, which may not be well captured by a
single global model. For this reason, the authors propose to cluster the time
series according to a vector of features extracted using the technique from
\citep{hyndman2015large} and the Snob clustering algorithm \citep{wallace2000mml}.
Only then, an RNN is trained per cluster,
after having decomposed the series into trend, seasonality and residual
components.  The RNN is followed by an affine neural layer to project the cell outputs to the dimension of the
intended forecast horizon. This approach is applied to publicly available datasets from time series
competitions, and appears to consistently improve against learning a single
global model. In subsequent work, \citet{bandara2019lstm} continued to mix heuristics, in this instance 
seasonality decomposition techniques, known from classical forecasting methods 
with standard  NN techniques. Their aim is to improve on scenarios with multiple 
seasonalities such as inter and intra daily. The findings are that for 
panels of somewhat unrelated time series, such decomposition techniques help global models
whereas for panels of related or homogeneous time series this may be harmful. 
The authors do not attempt to integrate these steps into the  NN architecture itself, which would allow for end-to-end learning. 

\citet{cinar2017position} proposed a content attention mechanism that seats on top of any seq2seq RNN. The idea is to select a combination of the hidden states from the history and combine them using a pseudo-period vector of weights to the predicted output step.

\citet{li2019enhancing} introduce two modifications to the Transformer architecture to improve its performance for forecasting. First, they include causal convolutions in the attention to make the key and query context dependent, which makes the model more sensitive to local contexts. Second, they introduce a sparse attention, meaning the model cannot attend to all points in the history, but only to selected points. Through exponentially increasing distances between these points, the memory complexity can be reduced from quadratic to $O(T (\log T)^2)$, where $T$ is the sequence length, which is important for long sequences that occur frequently in forecasting. 
Other architectural improvements to the Transformer model have also been used more recently to improve accuracy and computational complexity in forecasting applications. 
For example, \citet{lim2019temporal} introduce the Temporal Fusion Transformer (TFT), which incorporates novel model components for embedding static covariates, performing ``variable selection'', and gating components that skip over irrelevant parts of the context. The TFT is trained to predict forecast quantiles, and promotes forecast interpretability by modifying self-attention and learning input variable importance.
\citet{eisenach2020mqtransformer} propose MQ-Transformer, a Transformer architecture that employs novel attention mechanisms in the encoder and decoder separately, and consider learning positional embeddings from event indicators.
The authors discuss the improvements not only on forecast accuracy, but also on excess forecast volatility where their model improves over the state of the art.
Finally, \citet{zhou2020informer} recently proposed the Informer, a computationally efficient Transformer architecture, that specifically targets applications with long forecast horizons. 
The Informer introduces a {\em ProbSparse} attention layer and a distilling mechanism to reduce both the time complexity and memory usage of learning to $O(T \log T)$, while improving forecast performance over \deep forecasting benchmark.


\subsection{Deep State Space Models}\label{sec:state_space}

In contrast to pure deep learning methods for time series forecasting introduced in Section~\ref{sec:building_blocks}, \citet{rangapuram2018deep} propose to combine classical state space models (SSM)~\citep{durbin,hyndman08} with deep learning. 
The main motivation is to bridge the gap between SSMs that provide a principled framework for incorporating structural assumptions but fail to learn patterns across a collection of time series, and NNs that are capable of extracting higher order features but results in models that are hard to interpret.
Their method parametrizes a linear Gaussian SSM using an RNN. The parameters of the RNN are learned jointly from a dataset of raw time series and associated covariates. 
Instead of learning the SSM parameters $\theta_{i, 1:T_i}$ for the $i$-th time series individually or locally in the terminology of Section~\ref{ssec:formalization}), the model is global and learns a shared mapping from the covariates associated with each target time series to the parameters of a linear SSM. 
This mapping 
$\theta_{i,t} = f(\vxit{1:t}; \Phi)$, for $i = 1, \ldots, N$ and $t = 1, \ldots, T_i + h$,
is implemented by an RNN with weights $\Phi$ which are shared across different time series as well as different time steps.
Note that $f$ depends on the entire covariate time series up to time $t$ as well as the set of shared parameters $\Phi$. 
Since each individual time series $i$ is modelled using an SSM with parameters $\Theta_i$, assumptions such as temporal smoothness in the forecasts are easily enforced.
The shared model parameters $\Phi$ are learned by maximizing the likelihood given the observations $\mathcal{Z} = \{\vzit{1:T_i}\}_{i=1}^N$. 
The likelihood terms for each time series reduce to the standard likelihood computation under the linear-Gaussian SSM, which can be carried out efficiently via Kalman filtering~\citep{barber2012bayesian}.
Once the parameters $\Phi$ are learned, it is straightforward to obtain the forecast distribution via the SSM parameters $\theta_{i, T_i  + 1 : T_i + h}$. 

There are two major limitations of the method proposed in \citet{rangapuram2018deep}: first, the observations are assumed to follow a Gaussian distribution and second, the underlying latent process that generates observations is assumed to evolve linearly.
 \citet{de2020normalizing} address the first limitation via \textit{Normalizing Kalman Filters} (NKF) by augmenting SSMs with normalizing flows~\citep{realnvp, glow, TAN} thereby giving them the flexibility to model non-Gaussian, multimodal data.
Their main idea  is to map the non-Gaussian observations $\{\vzit{1:T_i}\}$ to more Gaussian-like observations via a sequence of \textit{learnable}, nonlinear transformations (e.g., a normalizing flow) so that the method in \citep{rangapuram2018deep} can then be applied on the transformed observations.
While being more flexible, their method still retains attractive properties of linear Gaussian SSMs, namely, tractability of exact inference and likelihood computation, efficient sampling, and robustness to noise.

In a concurrent work to~\citep{de2020normalizing}, ~\citet{kurle2020deep} improve the method in~\citep{rangapuram2018deep} by addressing both limitations.
In particular, to model nonlinear latent dynamics, they propose a recurrent switching Gaussian SSM, which uses additional latent variables to switch between different linear dynamics.
Moreover, to handle non-Gaussian observations, they propose a nonlinear emission model via a decoder-type NN~\citep{Fraccaro17}.
Although the exact inference is no longer tractable with these improvements, they show that the approximate inference and likelihood estimation can be Rao-Blackwellised; i.e., the inference for the Gaussian latent states can be done exactly while the inference for the switch variables needs to be approximated.

Finally, \citet{redsds} propose to extend \citep{rangapuram2018deep} via incorporating switching dynamics. The recurrent explicit duration switching dynamical system (RED-SDS) is a flexible model that is capable of identifying both state- and time-dependent switching dynamics of a time series. State-dependent switching is enabled by
a recurrent state-to-switch connection and an explicit duration count variable is
used to improve the time-dependent switching behavior. A hybrid algorithm that approximates the posterior
of the continuous states via an inference network and performs exact inference for
the discrete switches and counts provides efficient inference. The method is able to infer meaningful switching patterns
from the data and extrapolate the learned patterns into the forecast horizon.

\subsection{Multivariate Forecasting}\label{sec:multivariate}

The models presented up to this point are mainly global univariate models, i.e., they are trained on all time series but they are still used to predict a univariate target.
When dealing with multivariate time series, one should be able to exploit the dependency structure between the different time series in the panel in a generalization of Eq.~\eqref{eq:global_univariate} to Eq.~\eqref{eq:global_multivariate}. 

\citet{toubeau2018deep} and~\citet{salinas2019high} combined RNN-based models with copulas to model multivariate distributions. The model in~\citep{toubeau2018deep} uses a nonparametric copula to capture the multivariate dependence structure. In contrast, the work in~\citep{salinas2019high} uses a Gaussian copula process approach. \citet{salinas2019high} use a low-rank covariance matrix approximation to scale to thousands of dimensions. Additionally, the model implements a non-parametric transformation of the marginals to deal with varying scales in the dimensions and non-Gaussian data.
More recently, \citet{rasul2020multi} proposed to represent the data distribution with a type of normalizing flows called Masked Autoregressive Flows \citep{papamakarios2017masked} while using either an RNN or a Transformer \citep{vaswani2017attention} to model the  multivariate temporal dynamics of time series. Normalizing flows were also used to bring deep SSMs~\citep{rangapuram2018deep} to a flexible, multivariate scenario~\citep{de2020normalizing}. 
\citet{rasul2021autoregressive} propose TimeGrad which, like DeepAR, is an RNN model for which samples are drawn from the data distribution at each time step, with the difference that in TimeGrad the RNN conditions a diffusion probabilistic model \citep{sohl2015deep} which allows the model to easily scale to multivariate time series and accurately use the dependencies between dimensions.

A recent application of global multivariate models is for hierarchical forecasting problems~\citep{taieb2017coherent,wickramasuriya2015forecasting,taeib20,athanasopoulos2009hierarchical}. Typically, in such problems an aggregation structure is defined (e.g., via a product hierarchy) and a trade-off between forecast accuracy and forecast coherency with respect to the aggregation structure must be managed. Here, forecast coherency or consistency means that the forecasts conforms to the aggregation structure, so that aggregated forecasts are the same as forecasts of aggregated time series. This aggregation structure is typically encoded via linear constraints where the aggregation structure is captured in a matrix $S$. \citet{rangapur21} propose to use a multivariate model such as~\citep{salinas2019high} and enforce consistency of forecast samples via incorporation of a projection of the samples with $S$ into the learning problem. Dedicated work exists for aggregation along the time dimension~\citep{Theodosiou:2021a,ATHANASOPOULOS201760,rangapur22}.

In some multivariate forecasting settings the different dimensions are tied together by some interpretable connections other than a hierarchy and this can be modelled as part of the input layer rather than the output as discussed so far. One can for example think of forecasting the traffic network of a city where the traffic at each of the location in the city is mostly influenced by the traffic at the neighboring locations, like in PEMS-BAY and METR-LA \citep{li2018diffusion}.
Graph Neural Networks (GNN) have been used in this forecasting setting \citep{shang2021discrete, deng2021graph, wu2020connecting, kipf2018neural,garcia22,zuegner2021study} where, in addition to the forecasting task,  the challenge is to best use the graph information that is provided or even learn the graph if none is available.
The methods that propose to learn the graph do so by looking for the graph that allows to produce the most accurate forecasts.
An embedding is learned for each dimension, and similarity scores are computed between every two dimension using these embeddings from which the adjacency matrix is obtained, either by taking the K-top edges \citep{deng2021graph, wu2020connecting} or sampling from them \citep{shang2021discrete, kipf2018neural}.
As of now, two main strategies have been proposed to learn the node embeddings, either simply by gradient descent \citep{deng2021graph, wu2020connecting} or by taking representation from the time series \citep{shang2021discrete, kipf2018neural}, with the latter approach to seemingly yielding better results.
While these methods were all presented as point forecasting method, one could obtain probabilistic forecasts by training these models to parametrize a predictive distribution as explained in Section \ref{sec:output_transformation}.

\subsection{Physics-based Models}\label{sec:physics}

In \emph{physics-based} models, \deep forecasting methods have been proposed that model the underlying dynamics in sophisticated ways.  \citet{chen19} proposed the Neural ODE (NODE) model, where an ordinary differential equation (ODE) is solved forward in time, and the adjoint equation is solved backwards in time using backpropagation.  One limitation of the Neural ODE model is that the unknown parameters $\theta$ are assumed to be constant in time. Other limitations such as computational complexity have been addressed in follow-up work, e.g.,~\citep{marin22}. \citet{vialard20} extends the NODE model to allow the parameters $\theta(t)$ to be time-varying by introducing a shooting formulation. In the shooting formulation, the optimal $\theta$ is determined by minimizing a regularized loss function.  \citet{vialard20} also shows that a residual network (ResNet) can be expressed as the Forward Euler discretization of an ODE with time step $\Delta t= 1$.  \citet{wang_bridging_2020} compares successful time series deep sequence models, such as \citep{salinas2020deepar, rangapuram2018deep} to NODE and other hybrid deep learning models to model COVID-19 dynamics, as well as the population dynamics using the Lotka-Volterra equations.  Through their benchmarking study, the authors show that distribution shifts can pose problems for deep sequence models on these tasks, and propose a hybrid model AutoODE to model the underlying dynamics.

\subsection{\Gl}\label{sec:global_local}

With \textit{local models}, the free parameters of the model are learned individually for each series in a collection, see Section~\ref{ssec:formalization}.  Classical local time series models such as SSMs, ARIMA, and exponential smoothing (ETS) \citep{hyndman2018forecasting} excel at modelling the complex
dynamics of individual time series given a sufficiently long history.  Other local models include Gaussian SSMs, which are computationally efficient, e.g.,  via a Kalman filter, and Gaussian Processes (GPs) ~\citep{rasmussen2006gaussian,seeger2004gaussian, girard2003gaussian,brahim2004gaussian}.  These methods provide uncertainty estimates, which are critical for optimal downstream decision making.  Since these methods are local, they learn one model per time series and cannot effectively extract information across
multiple time series. These methods are unable to address cold-start problems where there is a need to generate predictions for a time series with little or no observed history.   

Conversely, recall that in \textit{global models}, their free parameters are learned jointly on every series in a collection of time series.  NNs have proven particularly well suited as global models~\citep{salinas2020deepar,gasthaus2019probabilistic,rangapuram2018deep, wen2017multi,laptev2017}.
Global methods can extract patterns from collections of irregular time series even when these patterns would not be distinguishable using a single series.

\Gl models have been proposed to combine the advantages of both global and local models.  Examples include mixed effect models ~\citep{crawley2012mixed}, which consist of two kinds of effects: fixed (global) effects that describe the whole population, and random (local) effects that capture the idiosyncratic of individuals or subgroups.  A similar mixed approach is used in hierarchical Bayesian~\citep{gelman2013bayesian} methods, which combine global and local models to jointly model a population of related statistical problems. In an early example of hierarchical Bayesian models, \cite{pmlr-v32-chapados14} combined global and local features for intermittent demand forecasting in retail planning. In~\citep{ahmed2012scalable,low2011multiple}, other combined global and local models are detailed.

A recent \gl family of models, Deep Factors~\citep{wang2019deepfactors} provide an alternative way to combine the expressive power of  NNs with the data efficiency and uncertainty estimation abilities of classical probabilistic local models.  Each time series, or its latent function for non-Gaussian data, is represented as the weighted sum of a global time series and a local model. The global part is given by a linear combination of a set of deep dynamic factors, where the loading is temporally determined by attentions. The local model is stochastic. Typical choices include white noise processes, linear dynamical systems, GPs \citep{maddix2018deep} or RNNs. The stochastic local component allows for the uncertainty to propagate forward in time, while the global NN model is capable of extracting complex nonlinear patterns across multiple time series. The \gl structure extracts complex nonlinear patterns globally while capturing individual random effects for each time series locally.  

The Deep Global Local Forecaster (DeepGLO) \citep{sen2019think} is a method that ``thinks globally and acts locally'' to forecast collections of up to millions of time series. It crucially relies on a type of temporal convolution (a so-called leveled network), that can be trained across a large amount time series with different scales without the need for normalization or rescaling.  DeepGLO is a hybrid model that uses a global matrix factorization model~\citep{yu2016} regularized by a temporal deep leveled network and a local temporal deep level network to capture patterns specific to each time series. Each time series is represented by a linear combination of $k$ basis time series, where $k \ll N$, with $N$ the total number of time series.  The global and local models are combined through data-driven attention for each time series.  

A further example in the \gl model class is the ES-RNN model proposed by \citet{smyl2018m4} that has recently attracted attention by winning the M4 competition \citep{makridakis2018m4} by a large margin on both evaluation settings. In the ES-RNN model, locally estimated level and trend components are multiplicatively combined with an RNN model. Apart from its \gl nature, it also integrates aspects of different model classes into a a single model similar to Deep State Space models (Section~\ref{sec:state_space}). In particular, the $h$-step ahead prediction $\hat{\mathbf{z}}_{i, t+1:t+h} = l_{i, t} \cdot \mathbf{s}_{i, t+1:t+h} \cdot \exp(\text{RNN}(\mathbf{x}_{i, t}))$ consists of a level $l_{i, t}$ and a seasonal component $s_{i, t}$ obtained through local exponential smoothing, and the output of a global RNN model $\text{RNN}(\mathbf{x}_{i, t})$, where $\mathbf{x}_{i, t}$ is a vector of preprocessed data extracted from deseasonalized and normalized time series $\mathbf{x}_{i, t} = \log(\mathbf{z}_{i, t-K:t}/(\mathbf{s}_{i, t-K:t} l_{i, t}))$ cut in a window of length $K+1$.
The RNN models are composed of dilated LSTM layers with additional residual connections. The M4-winning entry used slightly different architectures for the
different type of time series in the competition.

\subsection{Intermittent Time Series}\label{sec:intermittent}

We noted in the introduction that \deep forecasting models had a major impact on operational forecasting problems. In these large-scale problem, intermittent time series occur regularly~\citep{bose2017probabilistic}. Accordingly, research on 
 NNs for intermittent time series forecasting has been an active area. 
\citet{salinas2020deepar} propose a standard RNN architecture with a negative binomial 
likelihood to handle intermittent demand similar to~\citep{snyder12} in classical methods. 
To the best of our knowledge, other likelihoods that have been proposed for intermittent 
time series in classical models, e.g., by~\citep{seeger2016bayesian}, have not yet been carried over to  NNs. However, some initial work is available via more standard likelihoods~\citep{salinas2020deepar,JEON2021}.

In the seminal paper on intermittent demand forecasting \citep{Croston1972}, Croston separates the data in a sequence of observed non-zero demands and a sequence of time intervals between positive demand observations, and runs exponential smoothing separately on both series. 
A comparison of  NNs to classical models for intermittent demand first appeared in \citet{gutierrez_lumpy_2008}, where the authors compare the performance of a shallow and narrow MLP with Croston's method.
They find  NNs to outperform classical methods by a significant margin. 

\citet{kourentzes2013intermittent} proposes two MLP architectures for intermittent demand, taking demand sizes and intervals as inputs. 
As in \citet{gutierrez_lumpy_2008}, the networks are shallow and narrow by modern standards, with only a single hidden layer and three hidden units. 
The difference between the two architectures is in the output.
In one case interval times and non-zero occurrences are output separately, while in the other a ratio of the two is computed.
The approach proposed by \citet{kourentzes2013intermittent} outperforms other approaches primarily with respect 
to inventory metrics, but not forecasting accuracy metrics, challenging previous results in \citep{gutierrez_lumpy_2008}.
It is unclear whether the models are used as global or local. 
However, given the concern around overfitting and regularization, we assume that these models were primarily used as local models in the experiments. 

Both approaches of \citep{gutierrez_lumpy_2008,kourentzes2013intermittent} only offer point forecasts.
This shortcoming is addressed by~\citep{turkmen19,turkmen21}, where the authors propose renewal processes as natural models for intermittent demand forecasting.
Specifically, they use RNNs to modulate both discrete time and continuous time renewal processes, using the simple analogy that RNNs can replace exponential smoothing in~\citep{Croston1972}. 

Finally, a recent trend in sequence modelling employs NNs in modelling discrete event sequences observed in continuous time ~\citep{du_recurrent_2016,mei_neural_2017,xiao_wasserstein_2017,sharma2018point,turkmen2019fastpoint,shchur21} and~\citep{shchur2021neural} for an overview.
Notably, \citet{xiao2017joint} use two RNNs to parametrize a probabilistic ``point process'' model. 
These networks consume data from asynchronous event sequences and uniformly sampled time series observations respectively.
Their model can be used in forecasting tasks where time series data can be enriched with discrete event observations in continuous time. 

\subsection{Generalized Adversarial Networks}\label{sec:gans}

Additionally to the approaches mentioned in Sections~\ref{sec:output_transformation} and~\ref{ssec:further}, the recent literature contains further examples for density estimation, most prominently via Generalized Adversarial Networks (GANs)~\citep{goodfellow2014generative}. While GANs have received much attention in the overall deep learning literature~\citep{Karras_2019_CVPR, timegan_2019, engel2018gansynth, kevin2017, esteban2017}, this has not been reflected in forecasting. 
 We speculate that this is because a discriminator network can be replaced by metrics such as CRPS which measure the quality of generated samples. 
We therefore only provide a brief overview here and mention that, while they rely on the buildings blocks discussed in Section~\ref{sec:building_blocks}, they typically require architectures that are more complex than then ones discussed here and lead to involved optimization problems.

Despite the comparably less attention that GANs have received in forecasting, they have been recently applied to the time series domain~\citep{esteban2017, timegan_2019} to synthesize data~\citep{takahashi2019,esteban2017}
    or to employ an adversarial loss in forecasting tasks~\citep{wu2020}. Many time series GAN architectures use recurrent networks to model temporal dynamics~\citep{morgren2016, esteban2017, timegan_2019}. 
        Modelling long-range dependencies and scaling recurrent networks to higher lengths is inherently difficult and limits the application of time series GANs to short sequence lengths~\citep{timegan_2019, esteban2017}. 
        One way to achieve longer realistic synthetic time series is by employing convolutional~\citep{van2016wavenet,bai2018empirical,franceschi2020unsupervised} and self-attention architectures~\citep{vaswani2017attention}. 

Convolutional architectures are able to learn relevant features from the raw time series data~\citep{van2016wavenet,bai2018empirical,franceschi2020unsupervised}, but are ultimately limited to local receptive fields and can only capture long-range dependencies via many stacks of convolutional layers. 
Self-attention can bridge this gap and allow for modelling long-range dependencies from convolutional feature maps, which has been a successful approach in the image~\citep{zhang2019selfattention} and time series forecasting domain~\citep{li2019enhancing}. Another technique to achieve long sample sizes is progressive growing, which
    successively increases the resolution by adding layers to a GAN generator and discriminator during training~\citep{karras2018progressive}.
    A recent proposal~\citep{psagan} synthesizes progressive growing with convolutions and self-attention into a novel architecture particularly geared towards time series.

\subsection{Summary and Practical Guidelines}
\label{ssec:summary_models}

In Section~\ref{sec:building_blocks} and this section, we introduced a large number of \deep forecasting models. We summarize the main approaches in Table \ref{tab:model_summary}. The list below provide keys to reading the table. 

\begin{itemize}
	\item \emph{Forecast} distinguishes between probabilistic (\emph{Prob}) and \emph{Point} forecasts. 
	\item \emph{Horizon} indicates whether the model does one-step predictions (noted $1$) in which case multi-step forecasts are obtained recursively, or if it directly predicts a whole sequence ($\geq 1$).
	\item \emph{Loss} and \emph{Metrics} specifies the loss used for training and metrics used for evaluation. Here, we only provide an explanation of the acronyms and not the definition of each metric which can be easily found in the corresponding papers: negative log-likelihood (NLL), quantile loss (QL), continuous ranked probability score (CRPS), (normalized) (root) mean squared error (NRMSE, RMSE, MSE),  root relative squared error (RRSE), relative geometric RMSE (RGRMSE), weighted absolute percentage error (WAPE), normalized deviation (ND), mean absolute deviation (MAD), mean absolute error (MAE), mean relative error (MRE), (weighted) mean absolute percentage error (wMAPE, MAPE), mean absolute scaled error (MASE), overall weighted average (OWA), mean scaled interval score (MSIS), Kullback-Leibler divergence (KL), Value-at-Risk (VaR), expected shortfall (ES), empirical correlation coefficient (CORR), area under the receiver operating characteristic (AUROC), percentage best (PB).
\end{itemize}

While Table \ref{tab:model_summary} serves to illustrate the wealth of \deep forecasting methods now available, their sheer number may be slightly overwhelming. Furthermore, empirical evidence on the effectiveness of the different architectures has so far not revealed a clearly superior approach~\citep{alexandrov2019gluonts}. In this, forecasting differs from other domains, e.g., natural language processing where Transformer-based models~\citep{vaswani2017attention} dominate overall. Also, \deep forecasting methods seem to differ from other model families, such as tree-based methods where LightGBM~\citep{LightGBM17} or XGBoost~\citep{Chen2016} dominate (as in the recent M5 forecasting competition~\citep{januschowski21trees}). We speculate that this diffuse picture is in part due to the practical reasons, the relative immaturity of the field and the corresponding software implementations and in part due to fundamental reason as a natural consequence of the breadth and diversity of forecasting problems. 

So, choosing the appropriate architecture for a problem at hand can be a daunting task. In the following, we therefore attempt 
to provide guidelines for a more informed \deep forecasting model selection. These are largely based on our own experience in working 
with practical forecasting problem and they should primarily be taken as a non-exhaustive guidance on where to start model exploration.

\afterpage{%
    \clearpage
{\scriptsize
\begin{landscape}
\begin{longtable}{C{2cm}  C{2.5cm}  C{1.3cm}  C{1.0cm}  C{1.9cm}  C{1.9cm}  C{2.2cm}  C{3.1cm}}
	\toprule 
	Study & Structure & Forecast & Horizon & Loss & Metrics & Data Types & Comments \\
	\midrule
	\midrule
	\mbox{DeepAR} \citep{salinas2020deepar} & RNN & Prob & 1 & NLL & Coverage, QL, ND, NRMSE & demand, traffic, electricity & Learns parametric distributions\\
	\vspace{2mm}\\
	\citet{toubeau2018deep} & RNN/CNN & Prob & 1 & NLL/QL & RMSE, price & electricity & Nonparametric copula to capture multivariate dependence\\
	\vspace{2mm}\\
	\citet{salinas2019high} & RNN & Prob & 1 & NLL & QL, MSE & electricity, traffic, exchange rate, solar, taxi, wiki & Learns multivariate model via low-rank Gaussian copula processes\\
	\vspace{2mm}\\
	 \mbox{ARMDN} \citep{mukherjee2018armdn} & RNN & Prob & 1 & NLL & wMAPE & demand & Like \citep{salinas2020deepar}, but using mixture of Gaussian's and domain specific feature processing\\
	\vspace{2mm}\\
	\mbox{QARNN} \citep{xu2016quantile} & MLP & Prob & 1 & QL & VaR, ES & finance & Conditional quantile function estimation \\
	\vspace{2mm}\\
	\mbox{SQF-RNN} \citep{gasthaus2019probabilistic} & RNN & Prob & 1 & CRPS & QL, MSIS, NRMSE, OWA & demand, traffic, count data, finance, M4 & Models non-parametric distributions with splines\\
	\vspace{2mm}\\
	\mbox{LSTNet} \citep{lai2018modeling} & CNN + RNN + MLP & Point & 1 & $\ell_1$ & RRSE, CORR & traffic, solar, electricity, exchange rate & Extracts short and long temporal patterns with a CNN and RNN, respectively\\
	\vspace{2mm}\\
	\citet{zhu2017deep} & RNN + MLP & Prob & 1 & - & sMAPE, calibration &  daily trips &  Fits an encoder (RNN) that constructs an embedding state, which is fed to a prediction network (MLP)\\
	\vspace{2mm}\\
	\citet{laptev2017}  & RNN & Prob & 1 & MSE & sMAPE & traffic, M3 & LSTM as feature extractor\\
	\vspace{2mm}\\
	
	\citet{qiu2014ensemble} & MLP + SVR & Point & 1 & $\ell_2$ for MLP, SVR objective & RMSE, MAPE & energy, housing & Ensemble of DBNs where their output is fed to an SVR\\
	\vspace{2mm}\\		
	\mbox{A-LSTM} \citep{hsu2017time}  & RNN + MLP & Point & 1 & $\ell_2$, $\ell_2$ regularizer & RMSE & electricity consumption & Combination of LSTM with autoencoders\\
	\vspace{2mm}\\
	\citet{borovykh2017conditional}  & CNN & Point & 1 & $\ell_1$, $\ell_2$ regularizer & RMSE, MASE, HITS & index forecasting, exchange rate  & WaveNet \citep{van2016wavenet} based model adjusted for time series forecasting\\
	\vspace{2mm}\\	
	\mbox{SAnD} \citep{song2018attend}  & MLP + Attention & Point & 1 & $\ell_2$, cross-entropy, multi-label classification loss & AUROC, MASE, MSE & clinical & Transformer \citep{li2019enhancing} based model adjusted for time series forecasting\\
	\vspace{2mm}\\
	\citet{zhang2003time}  & MLP & Point & 1 & MSE & MSE, MAD & sunspot, lynx, exchange rate & Hybrid local model that uses ARIMA to capture the linear component and a NN for the nonlinear residuals\\
	\vspace{2mm}\\
	\citet{khashei2011novel}  & MLP & Point & 1 & MSE & MAE, MSE & sunspot, lynx, exchange rate & Hybrid local model that uses ARIMA and a NN for trend correction\\
	\vspace{2mm}\\
	\mbox{Deep State Space} \citep{rangapuram2018deep} & RNN + State Space & Prob & $\geq 1$ & NLL & P50, P90 quantile loss & traffic, electricity, tourism, M4 & RNN parametrized linear Gaussian SSM \\
	\vspace{2mm}\\
	\mbox{NKF} \citep{de2020normalizing} & RNN + State Space + Normalizing Flow (NF) & Prob & $\geq 1$ & NLL & QL & traffic, electricity, exchange rate, solar, wiki & RNN parametrized linear Gaussian SSM combined with normalizing flow, which acts as an emission model to handle non-Gaussian data\\
	\vspace{2mm}\\
	\mbox{ARSGLS} \citep{kurle2020deep} & Recurrent Switching State Space + NN & Prob & $\geq 1$ & NLL & QL & traffic, electricity, exchange rate, solar, wiki & Recurrent Switching State Space combined with decoder-type NN, which acts as an emission model to handle non-Gaussian data\\
	\vspace{2mm}\\
	\mbox{MQ-RNN/CNN} \citep{wen2017multi} & RNN/CNN + MLP & Prob & $\geq$1 & QL & QL, calibration, sharpness & demand & Learns pre-specified grid of quantiles \\
	\vspace{2mm}\\
	
	\citet{wen2019deep} & CNN + MLP & Prob &  $\geq$1 & QL, inverse reconstruction loss, NLL & QL, quantile crossing, QL over sum of future intervals & demand & Combines model in \citep{wen2017multi} with Gaussian copula \\
	\vspace{2mm}\\
	\mbox{DeepTCN} \citep{chen2019probabilistic} & CNN + MLP & Prob & $\geq$1 & QL & QL & retail demand & Learns pre-specified grid of quantiles \\
	\vspace{2mm}\\
	\mbox{N-BEATS} \citep{oreshkin2019n} & MLP & Point &  $\geq 1$ & sMAPE, MASE, MAPE & sMAPE, MASE, OWA & M4 & Deep, residual MLP that learns interpretable trend and seasonality function \\		
	\vspace{2mm}\\
	\citet{lv2014traffic} & Stacked autoencoder & Point & $\geq 1$ & MSE, KL sparsity constraint & MAE, MRE, RMSE & traffic & Stacked autoencoders with logistic regression output layer \\
	\vspace{2mm}\\
	\mbox{DCRNN} \citep{li2018diffusion} & RNN &  Point & $\geq 1$ & NLL & MAE, MAPE, RMSE & traffic &  Diffusion convolution for spatial and RNN for temporal dependencies\\
	\vspace{2mm}\\
	\citet{asadi2019spatial}  & CNN + RNN & Point & $\geq 1$ & $\ell_2$ & MAE, RMSE & traffic & Decomposition-based model for spatio-temporal forecasting\\	
	\vspace{2mm}\\	
	\citet{bandara2017forecasting}  & RNN + Classical Decomposition & Point & $\geq 1$ & - & sMAPE & CIF2016, NN5 & Clusters time series based on set of features and train one model per cluster\\
	\vspace{2mm}\\
	\mbox{LSTM-MSNet} \citep{bandara2019lstm} & RNN + Classical Decomposition & Point & $\geq 1$ & $\ell_1$ & sMAPE, MASE & M4, energy &  Decomposition based model with multiple seasonal patterns\\
	\vspace{2mm}\\
	\citet{cinar2017position} & RNN + Attention & Point & $\geq 1$ & $\ell_2$, $\ell_2$ regularizer & MSE, sMAPE & energy, max temperature, CPU usage, air quality  & Attention mechanism on top of RNN\\
	\vspace{2mm}\\	
			
	\mbox{Deep Factors} \citep{wang2019deepfactors} & RNN + GP & Prob & $\geq 1$ & NLL & QL, MAPE & electricity, traffic, taxi, uber & Global RNN and a local GP \\
	\vspace{2mm}\\
	\mbox{DeepGLO} \citep{sen2019think} & CNN & Point & $\geq 1$ & $\ell_2$ & WAPE, MAPE, sMAPE & electricity, traffic, wiki & Global matrix factorization regularized by a deep leveled network \\
	\vspace{2mm}\\
	\mbox{ES-RNN} \citep{smyl2018m4} & RNN & Point & $\geq 1$ & QL & MASE, sMAPE, MSIS & M4 & Locally estimated seasonality and trend and global RNN \\
	\vspace{2mm}\\
	\citet{kourentzes2013intermittent} & MLP & Point & 1 & $\ell_2$ & ME, MAE, service level & intermittent demand & MLP-based intermittent demand model \\
	\vspace{2mm}\\
	\mbox{Attentional Twin RNN} \citep{xiao2017joint} & RNN & Prob & 1 & NLL & MAE & point process data & Event sequence prediction\\
	\vspace{2mm}\\
	\citet{gutierrez_lumpy_2008} & MLP & Point & 1 & $\ell_2$ & MAPE, RGRMSE, PB & intermittent demand & MLP-based intermittent demand model \\
	\vspace{2mm}\\
	\mbox{Deep Renewal Process} \citep{turkmen19} & RNN & Prob & $\ge 1$ & NLL & P50, P90 quantile loss & intermittent demand & RNN-based intermittent demand model inspired by point processes \\
	\vspace{2mm}\\
	\mbox{WaveNet} \citep{van2016wavenet,alexandrov2019gluonts} & CNN & Prob & $\geq 1$ & NLL & mean opinion score & traffic, electricity, M4 & Diluted causal convolutions \\
	\vspace{2mm}\\
	\mbox{Transformer} \citep{li2019enhancing} & MLP & Point &  1 & NLL & QL & electricity, traffic, wind, M4, solar & Transformer with causal convolutions and sparse attention  \\	
	\vspace{2mm}\\
	 \mbox{AttnAR} \citep{yoo2021attention} & CNN & Point &  1 & $\ell_2$ & RMSE & electricity, traffic, solar, exchange rate & Multivariate forecasting  \\	
	 \vspace{2mm}\\
	 \mbox{LSTM MAF} \citep{rasul2020multi} & RNN & Prob &  $\geq 1$ & NLL & RMSE & electricity, traffic, solar, exchange rate & Multivariate forecasting using normalizing flows  \\	
	 \vspace{2mm}\\
	 \mbox{TimeGrad} \citep{rasul2021autoregressive} & RNN & Prob &  $\geq 1$ & NLL & RMSE & electricity, traffic, solar, exchange rate, taxi, wikipedia & Multivariate forecasting using diffusion models.  \\	
	 \mbox{TFT} \citep{lim2019temporal} & LSTM, MLP & Prob &  $\geq 1$ & QL & P50, P90 quantile loss & electricity, traffic, retail, volatility & Modified transformer architecture for improved interpretability  \\	
	 \vspace{2mm}\\
	 MQ-Transformer \citep{eisenach2020mqtransformer} & CNN, MLP & Prob & $\geq 1$ & QL & P50, P90, LT-SP & electricity, traffic, retail, volatility, retail demand (proprietary) & Architectural improvements on MQ-RNN/CNN for multi-step forecasting \\	
	 \vspace{2mm}\\
	 Informer \citep{zhou2020informer} & CNN, MLP & Point & $\geq 1$ & MSE & MSE, MAE & electricity, weather, sensor data & Sparse and computationally efficient transformer architecture  \\	
	 \vspace{2mm}\\
	\bottomrule \\	
\caption{Summary of modern \deep forecasting models.}
\label{tab:model_summary}
\end{longtable}
\end{landscape}
\clearpage
}
}

\subsubsection{Baseline methods and standard mode of deployment}
At the start of any in-depth model exploration, considering a baseline model is commonly accepted best practice. 
To the best of our knowledge, the most mature \deep forecasting models are DeepAR~\citep{salinas2020deepar} and MQCNN~\citep{wen2017multi} which exist in a number of  open-source and commercial implementations.\footnote{\url{https://aws.amazon.com/blogs/machine-learning/now-available-in-amazon-sagemaker-deepar-algorithm-for-more-accurate-time-series-forecasting/}} As a practical guideline, we recommend to start model 
exploration using at least these methods as baselines. Other candidates we would consider are N-BEATS~\citep{oreshkin2019n}, WaveNet~\citep{van2016wavenet} and a Transformer-based model. The relative performance of these methods compared with other methods should give reasonable, directional evidence whether the problem at hand is amenable to \deep forecasting methods.
We note that AutoML approaches for forecasting are available\footnote{\url{http://ai.googleblog.com/2020/12/using-automl-for-time-series-forecasting.html}} but while promising are in their infancy. At least in the M5 competition, they are still outperformed by the aforementioned more specialized \deep forecasting models.

Our typical suggestion is to employ NNs as global models since, given enough data, global
methods outperform classical local methods when dealing with groups of similar time series.\footnote{This is a more generally applicable fact beyond NN. 
 \citet{montero2020principles} show favorable theoretical and empirical properties for global over local models.} Interestingly, recent empirical evidence have shown that global models can achieve a state-of-the-art performance even in heterogeneous groups of time series. This is supported by the M4 \citep{makridakis2018m4} and M5 competitions where the top performing models had some form of globality. This suggests a more general applicability of global methods with a high impact on practical application where a general automated forecasting mechanism is required.

\subsubsection{Data characteristics}
The amount of data available is among the easiest dimensions in choosing a \deep forecasting model. First, NNs require a minimum amount of data to be effective in comparison to other, more parsimoniously parametrized models. This is perhaps the most important factor in successful applications of NNs in forecasting. 
How much data does one need for a given application? Several important points should be discussed on this question. First, the amount of data is often misunderstood as the \emph{number of time series} but in reality the amount of data typically relates to the \emph{number of observations}. For instance, one may have only one time series but many thousands of observations, as in the case of a time series from a real-time sensor where measurements happen every second for a year, allowing to fit a complex NN \citep{anomaly}. Second, it is probably better to see the amount of data in terms of \emph{information quantity}. For instance, in finance 
the amount of information of many millions of hourly transactions is limited
given the very low signal-to-noise ratio in contrast to a retailer whose products follow clear seasonality and patterns, making it easier to apply deep learning methods. The more structured the data is (e.g., via strong seasonality or knowledge about the underlying process) the better \deep forecasting models that incorporate these structures will fare. On the contrary, if the time series are more irregular or short, a more data-driven approach (e.g., via Transformer-based models) will often be preferable. The importance of covariate information for the forecasting problem at hand can further help determine the correct method. Some NN architectures need extensions to include such information while others readily accept them.

From a practical perspective,  NNs have been reported to outperform demand forecasting baselines starting from 50000 observations in \citep{salinas2020deepar} and from a few hundred observations in load-forecasting \citep{rangapuram2018deep,wang2019deepfactors}.  Understanding better these limitation, both theoretically and empirically, is an area of current research and is not yet well understood. See \citep{kuznetsov} for some current theoretical work on sample complexity of \gl approaches for instance and~\citep{borchert2022multiobjective} for empiricial work.

\subsubsection{Problem characteristics}
The characteristics of the forecasting problem to be solved are natural important decision points. We list a few dimensions to consider here. 

One important aspect of a model is its forecast nature, i.e., if it produces probabilistic or point forecasts. The choice of this is highly dependent on the underlying application. To illustrate this we can examine two different forecasting use cases: product demand and CPU utilization. In the former use case one wishes to forecast the future demand of a product in order to take a more informed decision about the stock that is required to have in a warehouse or to optimize the labour planning based on the traffic that is expected. In the latter, the forecast of CPU utilization could be used to identify in a timely manner if a process will fail in order to proactively resolve associated issues, or to detect possible anomalous behaviours that could trigger some root cause analysis and system improvements. Although in both applications a forecast is required, the end goal is different, which changes the requirements of the chosen forecasting model. For example, for product demand the whole distribution of the future demand might be important: one cannot rely on a single forecast value since the variance in the forecast plays an important role to avoid out of stock issues or under/over planning the expected required labour. Therefore, in this application it is important to use a model that focuses on predicting accurately the whole distribution. On the other hand, for CPU utilization one might be interested in the $99$-th percentile, since everything below that threshold might not be of particular interest or does not produce any actionable alarm. In this case, a model that focuses on a particular quantile of importance is of higher interest than a model that predicts the whole distribution with possibly worse accuracy on the selected quantile.

It is observed~\citep{salinas2020deepar, rangapuram2018deep, salinas2019high, de2020normalizing, kurle2020deep} empirically that autoregressive models are superior in performance (in terms of forecast accuracy) compared to state space models, especially when the data is less noisy and the forecast horizon is not too long.
This is not surprising given that the autoregressive models directly use past observations as input features and treat own predictions as lag inputs in the multi-step forecast setting. A general rule of thumb is that if one knows details such as the forecast horizon, the quantile to query or the exact goals of the forecasting problem in advance and these are unlikely to change, then a discriminative model is often a good default choice.
Conversely, state space models proved to be robust when there are missing and/or noisy observations~\citep{de2020normalizing}.
Moreover, if the application-specific constraints can be incorporated in the latent state, then state space models usually perform better even in the low-data regimes~\citep{rangapuram2018deep}.

The length of the forecast horizon relative to the history or, more generally speaking, the importance of the historic values for future values must further be taken into account. For example, very long forecast horizons may require to control (e.g., via differential equations) the exponential growth in the target. A canonical example for this is forecasting of a pandemic. This example further 
clarifies the importance of being able to produce counterfactuals for what-if analysis (e.g., the incorporation of intervention). Not all \deep forecasting models allow for this.

\subsubsection{Other Aspects}
A number of other aspects can further help to narrow the model exploration space. For example, computational constraints (how much time/money for training is available, are there constraints on the latency during inference) can favor ``simpler'' NNs, see e.g.,~\citep{borchert2022multiobjective} for a discussion on multi-objective forecasting model selection. Another aspect to consider could be CNN over RNN-based architectures. The skill set of the research team available is an important factor. For example, probabilistic models often are more sensitive towards parametrization and identifying reasonable parameter ranges requires in-depth knowledge. On the other extreme, troubleshooting Transformer-based models requires deep learning experience that not every research team may possess. The time budget available for model development and the willingness to extend existing models are further factors.


\section{Conclusions and Avenues for Future Work}
\label{sec:conclusions}

This article has attempted to provide an introduction to and an overview of NNs for forecasting or \deep forecasting. We began by providing a panorama of some of the core concepts in the modern literature on NNs chosen by their degree of relevance for forecasting. We then reviewed the literature on recent advances in \deep forecasting models.

\Deep forecasting methods have received considerable attention in the literature because they excel at addressing forecasting problems with many related time series and at extracting weak signals and complex patterns from large amounts of data. From a practical perspective, the availability of efficient programming frameworks helps to alleviate many of the pain points that practitioners experience with other forecasting methods such as manual feature engineering or the need to derive gradients. 
However, NNs are not a silver bullet. For many important classes of forecasting problems such as long-range macro-economic forecasts or other problems requiring external domain knowledge not learnable from the data, \deep forecasting methods are not the most appropriate choice and will likely never be. Still, it is our firm belief that NNs belong to the toolbox of every forecaster, in industry and academia.

Building onto the existing promising work in NNs for forecasting, many challenges remain to be solved. We expect that the current trends of hybridizing existing time series techniques with NNs~\citep{rangapuram2018deep,smyl2018m4,gasthaus2019probabilistic,turkmen19} and bringing innovations from other related areas or general purpose techniques to forecasting~\citep{goodfellow2014generative,van2016wavenet,vaswani2017attention} will continue organically. 
Typical general challenges for  NNs, such as data effectiveness, are important in forecasting and likely need a special treatment (see \citep{FawazFWIM18} for an approach in time series classification with transfer learning). Other topics of general ML interest such as interpretability, explainability and causality (e.g.,~\citep{binder16,Lipton:2018,schlkopf2019causality}) are of particular practical importance in the forecasting setting. It is our hope that original methods such as new  NN architectures will be pioneered in the time series prediction sector (e.g., \citep{oreshkin2019n}) and that those will then feed back into the general NN literature to help solve problems in other disciplines. 

Beyond such organic improvements, we speculate that another area in which NNs have had tremendous impact~\citep{AlphaGo,Silver1140} may become important for forecasting, namely deep reinforcement learning. In contrast to current practice, where forecasting merely 
serves as input to downstream decision problems (often mixed-integer nonlinear stochastic optimization problems), for example to address problems such as restocking decisions, reinforcement learning allows to directly learn optimal decisions in business context~\citep{januschowski18}. It will be interesting to see whether reinforcement based approaches can improve decision making -- and how good forecasting models could help improve reinforcement approaches.

As methodology advances, so will the applicability. Many potential applications of forecasting methods are under-explored. To pick areas that are close to the authors' interests, in database management, cloud computing, and system operations a host of applications would greatly benefit from the use of principled forecasting methods (see e.g.,~\citep{resilient20,ayed2020anomaly,flunkert2020simple}). 
Forecasting can also be used to improve core ML tasks such as hyperparameter optimization (e.g.,~\citep{Domhan2015}) and we expect more applications to open up in this area.

\bibliography{tutorial_bibliography}

\end{document}